\documentclass[letterpaper, 10 pt, conference]{ieeeconf}  

\IEEEoverridecommandlockouts                              

\usepackage{xcolor}
\usepackage{amsmath, amsfonts}
\usepackage{graphicx}
\usepackage{tabularray}
\usepackage{subcaption} 
\usepackage{caption}    
\usepackage{cite}
\usepackage{balance}

\title{\LARGE \bf
Dexterous Control of an 11-DOF Redundant Robot for CT-Guided Needle Insertion With Task-Oriented Weighted Policies$^\star$
}

\author{
Peihan Zhang$^{1}$,
Derek Chen$^{1}$,
Ishan Duriseti$^{1}$,
Florian Richter$^{1}$,
Zoe Chiu$^{2}$,
Moira Bohley$^{1}$,\\
Albert Hsiao$^{3}$,
Sean Tutton$^{3}$,
Alexander Norbash$^{4}$,
Michael Yip$^{1}$, \IEEEmembership{Senior Member, IEEE}
\thanks{$^{\star}$This work was funded by the National Institutes of Health Award 1R01CA278703-01. The authors would like to thank Stäubli corporation for their assistance and support of this project.}%
\thanks{$^{1}$Jacobs School of Engineering, University of California San Diego, La Jolla, CA 92093 USA. {\tt\footnotesize\{pez004, dec004, iduriseti, frichter, dbohley, m1yip\}@ucsd.edu}}%
\thanks{$^2$School of Electrical and Computer Engineering, Cornell University, Ithaca, NY 14853 USA. {\tt\footnotesize zc499@cornell.edu}}%
\thanks{$^3$School of Medicine, University of California San Diego, La Jolla, CA 92093 USA. {\tt\footnotesize\{a3hsiao, stutton\}@ucsd.edu}}%
\thanks{$^4$School of Medicine, University of Missouri-Kansas City, Kansas City, MO 64110 USA. {\tt\footnotesize\{norbash\}@umkc.edu}}%
}

\begin{document}

\maketitle
\thispagestyle{empty}
\pagestyle{empty}

\begin{abstract}
Computed tomography (CT)-guided needle biopsies are critical for diagnosing a range of conditions, including lung cancer, but present challenges such as limited in-bore space, prolonged procedure times, and radiation exposure. Robotic assistance offers a promising solution by improving needle trajectory accuracy, reducing radiation exposure, and enabling real-time adjustments.
In our previous work, we introduced a robotic platform designed for accurate needle insertion within the confined CT bore. However, its performance in clinical settings is restricted by limited dexterity and a constrained workspace.
In this study, we present an 11-degree-of-freedom (DOF) robotic system that integrates a 6-DOF robotic base with an improved 5-DOF cable-driven end-effector, yielding a significantly expanded workspace and enhanced dexterity.
To leverage the hyper-redundant degrees of freedom, we introduce a weighted inverse kinematics controller, along with a null-space control strategy to optimize maneuverability and dexterity.
By using a task-oriented weight matrix as a hyperparameter, the system provides a two-stage priority scheme fit for both large-scale movement and fine in-bore adjustments.
In clinically relevant simulated scenarios, the system demonstrates a consistent 97\% reachability rate across various human models.
In addition, the task-oriented weight-matrix policy is extensively explored in five representative subtasks seen during needle biopsy through both simulation and real-world experiments, demonstrating superior tracking accuracy and enhanced manipulability for CT-guided procedures.

\end{abstract}

\section{INTRODUCTION}

Lung cancer is the most frequently diagnosed type of cancer, accounting for one in eight cases worldwide, and remains the leading cause of cancer-related deaths~\cite{chhikara2023global}. Seventy percent of lung cancer tumors are in advanced and unresectable stages, necessitating the use of needle biopsy as a tool in the early detection process~\cite{travis2013diagnosis, wu2011ct}. Computed tomography-guided needle biopsy (CT-NB) is currently the most common minimally invasive procedure for crucial diagnosis and treatment for such a disease~\cite{saifuddin2021current}.
 
Traditionally, interventional radiologists (IRs) perform CT-guided procedures to locate lesions and insert a needle-shaped probe into the target. This process involves a repetitive, multi-step workflow in which IRs alternate between withdrawing the patient from the CT scanner bore to manually and incrementally advance the needle, and stepping away from the gantry and re-scanning the patient to update the needle and lesion position \cite{guiu2021feasibility}.
IRs must balance the frequency of scans which assess tip-to-target accuracy against the risk of radiation exposure for both themselves and the patient~\cite{leng2018radiation}. The delayed feedback of CT scanning and freehand adjustments results in prolonged procedures, multiple insertion attempts~\cite{chiu2021costs}, frequent complications~\cite{zhang2020biopsy} and side effects~\cite{wiener2011population}. 

Robotic platforms offer a promising alternative to manual CT procedures by increasing needle trajectory accuracy, reducing radiation exposure, and shortening procedure time by enabling precise tool adjustments while the patient remains within the bore~\cite{lanza2023robotics}. 
Needle insertion robotic platforms targeting the torso and chest regions face many challenges, in part because the chest and abdomen constitute the largest portion of the anatomy, limiting the room available for robotic systems within the confined imaging bore~\cite{schreiber2022crane}. Moreover, procedures in this region must contend with numerous anatomical obstacles, including the rib cage and major blood vessels, as well as substantial anatomical motion from lung expansion and vascular pulsation. Consequently, robotic platforms employed for torso and chest procedures need to offer high dexterity, exceptional accuracy, a low-profile design, and a sufficiently large working range within the CT imaging bore~\cite{schreiber2024crane}.

\subsection{Related Works}

\begin{figure}[t]
    \centering
    \includegraphics[width=\linewidth]{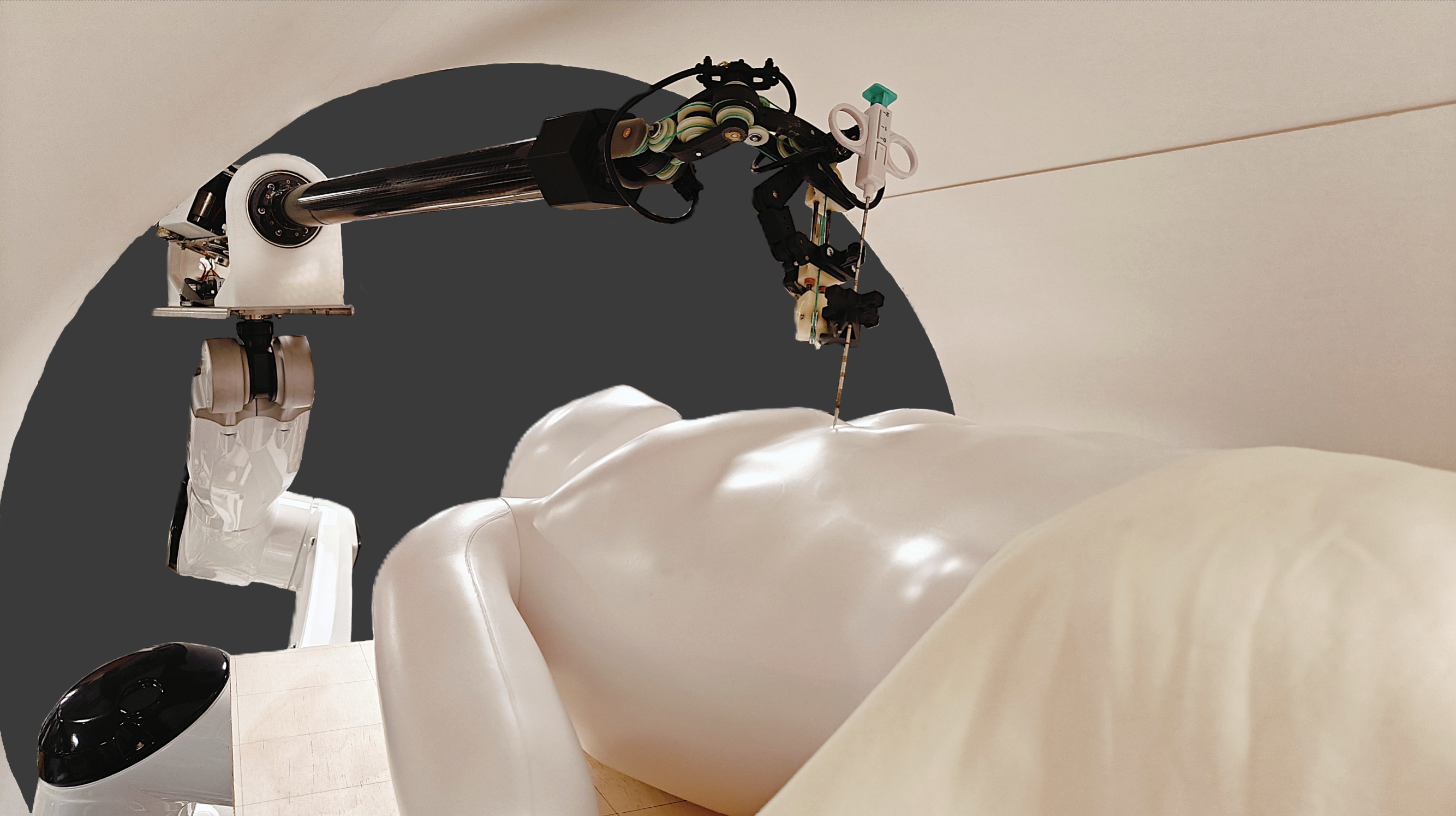}
    \caption{
    Dexterous 11-DOF redundant robotic system executes needle insertion task in confined image bore.
    }
    \label{fig:sys-cover}
\end{figure}

Many robotic platforms have been developed for operation within a CT bore, varying in size, needle insertion methods, and the extent of human involvement required\cite{siepel2021needle, schreiber2024crane}. Needle insertion platforms can be broadly categorized as either passive or fully active. Typically, passive systems are patient-mounted~\cite{levy2021clinical, gunderman2023autonomous, hungr2016design} or couch-mounted~\cite{schreiber2019open, schulz2013accuracy}, so the system translates together with the patient/couch during intraoperative scanning. Some designs even leverage the inherent motion that accompanies patient respiration~\cite{gunderman2023autonomous}.
However, passive systems require the physician to manipulate the device into the correct position by hand, which precludes fully automatic setup and increases both time and complexity.
Furthermore, patient-mounted systems must balance system stability with patient comfort, whereas couch-mounted systems risk damage to the device and potential patient injury in the event of improper fixation.

Meanwhile, fully active systems are typically floor-mounted, offering a more straightforward procedure setup, a large active workspace, and compatibility with remote teleoperation~\cite{komaki2020robotic, won2017validation, spenkelink2023evaluation}.
Conversely, floor-mounted systems have to compensate for the translation of the couch during intraoperative scanning to ensure patient safety~\cite{kim2017impedance}.
It is common to use industrial robots with a custom end effector in a floor-mounted scheme~\cite{spenkelink2023evaluation, chen2023robot, lei2023robotic}. These designs benefit from the high stiffness and accuracy of industrial robots but incur a larger system size, reduced in-bore dexterity, and a more limited workspace. However, when carefully designed, the arm+manipulator configuration can be advantageous compared to a single-manipulator system.

In our previous work~\cite{schreiber2024crane, schreiber2022crane}, we introduced a redundant robotic platform with high dexterity and accuracy for CT-guided needle insertion. This platform features a fully active workspace and requires no manual device setup.
Its cable-driven design isolates the drive motors from the end-effector, enabling a low profile within the scanner bore, minimal backlash, low inertia, and backdrivability. 
However, the system’s 3-DOF base was cumbersome and relied on serial linear motion, restricting deployment in real clinical applications and constraining its dexterity. The previous system achieved nearly 80\% in-bore reachability, indicating insufficient dexterity to handle certain corner cases.


\subsection{Contributions}

In this paper, we build on our previous work ~\cite{schreiber2024crane, schreiber2022crane} to perform percutaneous CT-guided procedures. Our key contributions include:
1) Implementation of a new redundant 11-DOF needle biopsy platform; the 5-DOF end-effector is significantly updated mechanically and electronically for integration with a 6-DOF arm.
2) Evaluation of the platform’s larger active workspace and enhanced in-bore reachability in clinically relevant simulations; the system achieves $>$97\% reachability across various human models.
3) Exploration of task-oriented weight policy selection with a weighted inverse kinematics controller and null-space subgoals. Both large-scale movements into the imaging bore and fine control for needle positioning are satisfied by incorporating a two-stage user-defined weight matrix.
The study provides a strong case for hyper-redundancy and null-space control formulations for robot-assisted needle biopsy scenarios. We also provide a demonstration of teleoperated control to show the intended workflow of our system, leveraging the 6-DOF robotic base for large movements and the 5-DOF end-effector for fine control for needle positioning.

\section{METHOD}
This section first presents the new mechanical-electrical design and software architecture of the 11-DOF system. A weighted inverse kinematics controller is then introduced to leverage the system's dexterity and large workspace. Finally, the robot's redundancy enables the use of a null-space controller to further improve dexterity during the needle biopsy task.

\subsection{System Overview}

\begin{figure}[t]
    \centering
    \includegraphics[width=\linewidth]{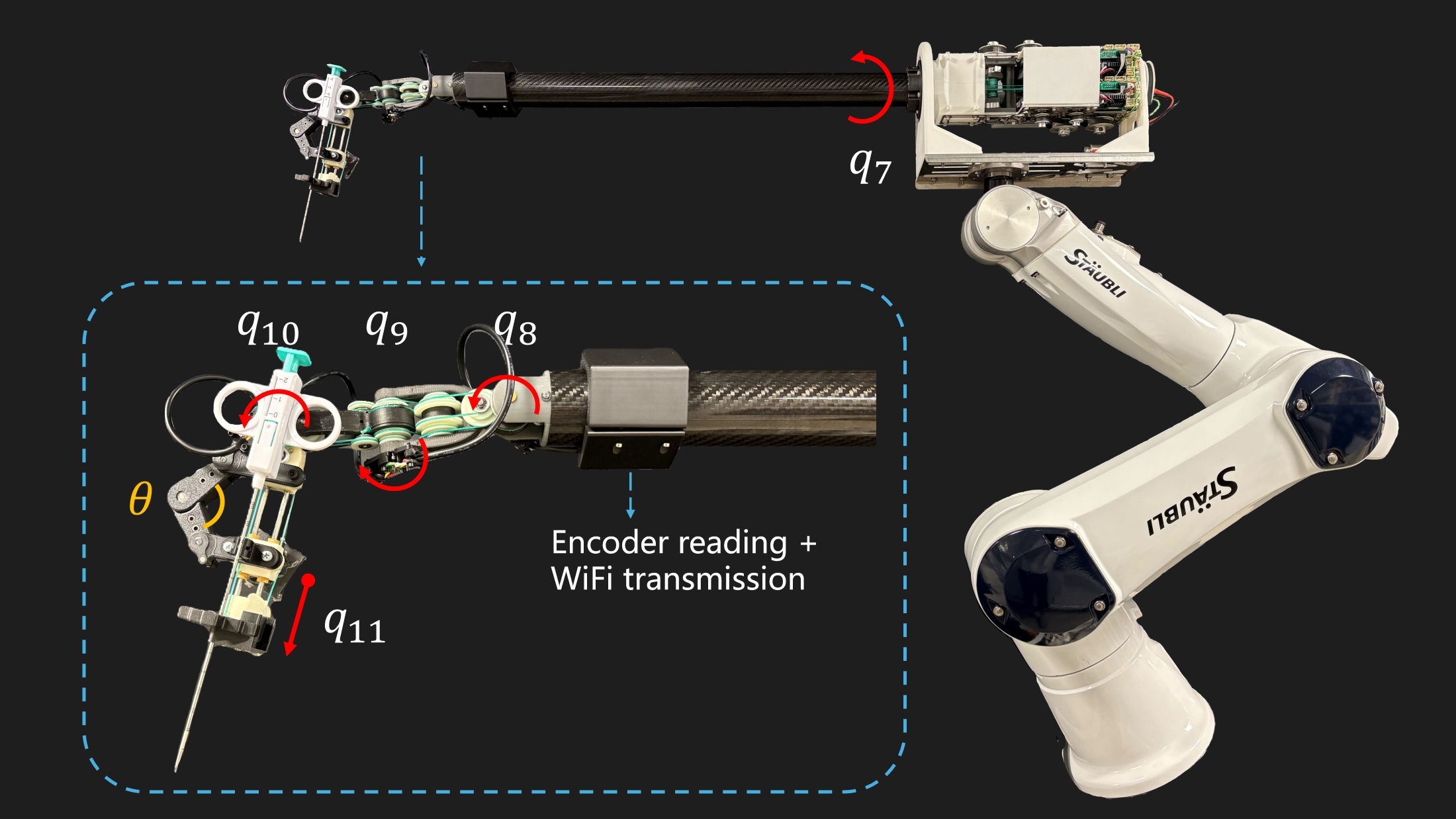}
    \caption{Our dexterous 11-DOF redundant robotic system utilizes a 6-DOF robotic base, while a 5-DOF cable-driven end-effector provides precise needle positioning.
        A translate-to-rotate mechanism is used to measure the 11th joint. 
        The end-effector's joint positions are transmitted wirelessly.
    }
    \label{fig:sys-real-sim}
\end{figure}

Building on our previous work, the new 11-DOF redundant robot consists of a 6-DOF robotic arm and an upgraded 5-DOF cable-driven end-effector.
The arm supplies large-scale positioning and reorientation around the scanner bore, while the end-effector provides local precision for needle alignment and insertion.
\subsubsection{Mechanical and Electrical Design}

For fine manipulation, we significantly refined the 5-DOF cable-driven end-effector developed in previous work~\cite{schreiber2024crane, schreiber2022crane}. The original design excelled at dexterous, fine-needle manipulation within confined spaces due to its compact, low-inertia structure. However, it also had some practical drawbacks, including being relatively heavy and cumbersome to set up, while having complex wiring that restricted joint movement. Our new version addresses these issues through extensive mechanical and electrical redesign. The upgrades are summarized in Table \ref{tab:mech}.

\begin{table}
\centering
\caption{{System upgrades of the new 11-DOF system}}
\label{tab:mech}
\begin{tblr}{
  width = \linewidth,
  colspec = {Q[60]Q[300]Q[300]Q[300]},
  cells = {c},
  cell{1}{1} = {c=2}{},
  cell{2}{1} = {r=5}{},
  cell{7}{1} = {r=2}{},
  vline{3} = {1-8}{},
  hline{1-2,7,9} = {-}{},
}
system              &                              & 11-DOF (our)           & 8-DOF (previous)    \\
{End\\Effector    } & Weight                  & 3.1 kg                        & 4.6 kg               \\
                    & Base length             & 0.31 m                             & 0.43 m                  \\
                    & Main axis range              & $[-2 \pi, 2\pi]$              & $[-\pi/2, \pi/2]$                  \\
                    & Joint data reading           & Wireless                      & Wired                \\
                    & Insertion measure      & Magn. encoder & Magn. tracker                     \\
Base                & Type                         & 6-DOF arm                     & 3-DOF linear~        \\
                    & Range                        & 920 mm max                       & 400 mm XYZ
\end{tblr}
\end{table}

Mechanically, the end-effector is improved to reduce distal mass and shorten motor spacing while preserving stiffness, facilitating integration with the 6-DOF arm.
The total weight is reduced by approximately 33\%, from 4.6 kg to 3.1 kg.
The length of the base that holds the motors and controllers is shortened by 28\% to 0.31 m.
We also updated the measurement method for the needle-insertion joint (the system’s 11th DOF) by engineering a compact translation-to-rotation mechanism, as shown in Fig.~\ref{fig:sys-real-sim}.
This mechanism converts the linear motion of the needle driver into rotational motion. The linear displacement of the needle—i.e., the joint position of the system’s 11th DOF, $q_{11}$—is computed directly from the encoder’s measured angle $\theta$ (in radians) using the relationship:
$$q_{11} = 2 l \sin(\theta/2)$$
where $l$ is the measurement-arm length.
Since the magnetic encoder has $<$ 0.1$^\circ$ angle error, the resulting linear error is $<$ 0.1 mm, comparable to the magnetic tracker used in the previous system.

Electrically, leveraging the demonstrated electromagnetic compatibility of wireless transmission in a computed tomography environment \cite{li2018anglenav}, we eliminated the long, cumbersome data cable.
All joint magnetic encoders are daisy-chained via an I\textsuperscript{2}C bus to an onboard microprocessor.
The module is powered by a battery, and sensor data is transmitted wirelessly to the main controller via Wi-Fi.
This self-contained wireless design drastically simplifies system setup.
With wireless transmission and unified sensing, the rotational limit of the end-effector’s main-axis joint (the 7th DOF) increased from $\pm \pi/2$ to $\pm 2\pi$, improving the system's dexterity.

Since the previous 8-DOF system was mechanically cumbersome and its 3-DOF serial linear base constrained reachability to “in-front-of-the-scanner” configurations, we replaced the base with a 6-DOF commercial arm (Stäubli TX2-60L).
The new arm provides higher dexterity and expands the robot’s maximum reach, enabling flexible large-scale positioning of the end-effector during biopsy procedures.
A detailed analysis of the new system’s workspace and dexterity is provided in the subsequent section.

\subsubsection{Software Architecture}
\begin{figure}[t]
    \centering
    \includegraphics[width=0.95\linewidth]{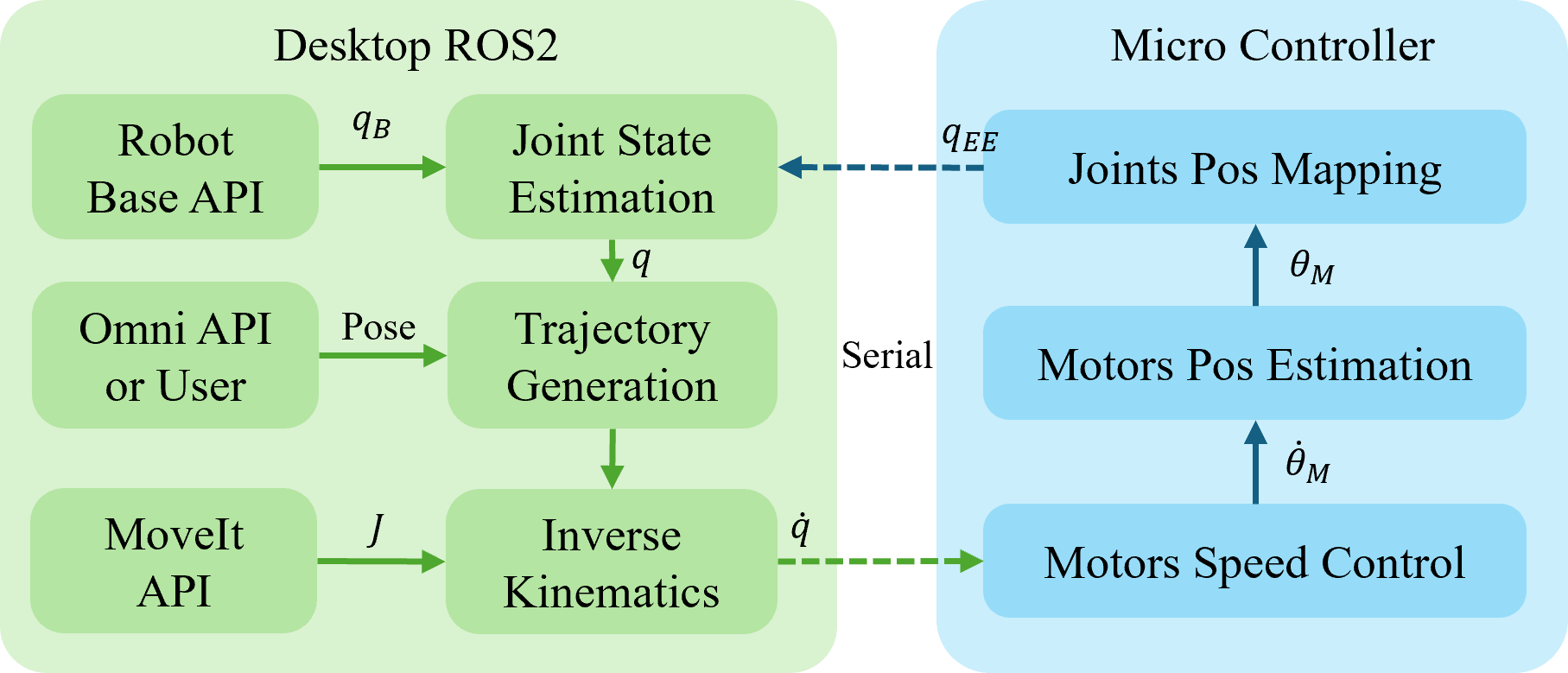}
    \caption{The software architecture consists of high-level control and planning modules running on a desktop in ROS2 framework, while low-level real-time controllers operate on embedded microcontrollers.}
    \label{fig:software-arch}
\end{figure}

The software architecture, as illustrated in Fig. \ref{fig:software-arch}, is divided into high-level control and planning modules running on a desktop computer and low-level real-time embedded microcontrollers. The desired needle pose is provided either by a 3D stylus (Phantom Omni) or through user commands and is then compared with the current end-effector pose. A joint velocity command is generated by the inverse kinematics solver and transmitted to the microcontroller via serial communication. The microcontroller records motor speed, estimates joint positions using the coupling matrix, and provides joint state feedback to the desktop-based state estimator. The details of the coupling matrix are given in \cite{schreiber2024crane}; broadly speaking, it is used to decouple joints that move together when a cable-drive is used in serial-link kinematics.

The new 11-DOF robotic system introduces challenges and opportunities in robot control, which are addressed in the following subsections by focusing on controlling large-scale movement and fine in-bore adjustments.

\subsection{Weighted Inverse Kinematics Controller}
A weighted inverse kinematics controller was developed for our redundant 11-DOF robot, enabling the accurate tracking of the desired needle pose while allowing users to dexterously manipulate the needle within the scanner.

Consider a robot Jacobian $J(q)\in\mathbb{R}^{m\times n}$ with $n$ joint DOFs and $m$ task DOFs. In our application, $n=11$ as the last DOF of the robot is a prismatic joint for insertion and kept static during target tracking state. The task error is represented in $m=6$ dimensions (position and orientation), but rotation about the needle's principal axis need not be controlled, so the task Jacobian has rank $5$ and only five directions are independently constrained.

Since the joint DOFs are strictly greater than the task space DOFs, there are infinity many solutions to the inverse kinematics problem. Although a robust pseudo-inverse or a singular value decomposition scheme is often taken to form the inverse, these do not leverage task-level information. Instead, we propose prioritizing joint movements based on the characteristics of the robot platform and task requirement. To achieve this, we introduce a positive-definite diagonal weight matrix $\mathcal{W}=\mathrm{diag}(w_1,\dots,w_n)$ to penalize joint motions, where a larger $w_i$ implies a higher penalty for moving joint $i$.
As a result, we formulate the problem as seeking joint velocity $\dot{q}$ that minimizes a cost function combining task error and a \textit{weighted} joint effort:
\begin{equation}
\label{eq:cost-func}
\min _{\dot{q}} \frac{1}{2}\left\|J \dot{q}-\dot{x}_{\mathrm{des}}\right\|^2+\frac{\lambda^2}{2} \dot{q}^T \mathcal{W} \dot{q}
\end{equation}
where $\dot{x}_{\text{des}}\in\mathbb{R}^6$ is the generalized desired end-effector velocity and $\lambda>0$ is a small damping factor. The first term is the task-space error, and the second term penalizes joint velocities (weighted by $\mathcal{W}$). 
Taking the derivative of \eqref{eq:cost-func} with respect to $\dot{q}$ and setting it to zero yields
\begin{equation}
\label{eq:J-inv-proof-1}
J^T\left(J \dot{q}-\dot{x}_{\mathrm{des}}\right)+\lambda^2 \mathcal{W} \dot{q}=0 .
\end{equation}
Solving \eqref{eq:J-inv-proof-1} gives 
\begin{equation}
\label{eq:J-inv-proof-3}
\dot{q}=\left(J^T J+\lambda^2 \mathcal{W}\right)^{-1} J^T \dot{x}_{\mathrm{des}}
\end{equation}
which is the damped weighted least-squares solution. Through a change of variables, we can further rearrange \eqref{eq:J-inv-proof-3} as 
\begin{equation}
\label{eq:J-inv-proof-4}
\dot{q}=\mathcal{W}^{-1} J^T\left(J \mathcal{W}^{-1} J^T+\lambda^2 I_m\right)^{-1} \dot{x}_{\mathrm{des}}
\end{equation}
where $I_m \in \mathbb{R}^{m\times m}$ is the identity matrix. 
From \eqref{eq:J-inv-proof-4}, we define the weighted damped pseudo-inverse as
\begin{equation}
\label{eq:robust-weight-inv-def}
J_{\mathcal{W}}^{\dagger}:=\mathcal{W}^{-1} J^T\left(J \mathcal{W}^{-1} J^T+\lambda^2 I_m\right)^{-1}
\end{equation}
As a result, we obtain
\begin{equation}
\label{eq:robust-weight-inv}
\dot{q} = J_{\mathcal{W}}^{\dagger} \dot{x}_{\text{des}}
\end{equation}

Note that if $\mathcal{W}$ becomes an identity matrix (representing equal joint weights), then \eqref{eq:robust-weight-inv-def} reduces to the robust pseudo-inverse $J^{\dagger} = J^T(JJ^T + \lambda^2 I)^{-1}$ from the Levenberg–Marquardt algorithm.
Assigning different joint weights prioritizes which joints contribute to an end-effector motion, thereby affecting performance metrics such as trajectory-tracking accuracy and response speed. The effects of different weight policies on system performance are explored and analyzed in Section \ref{sec:exp-weight}.

To apply \eqref{eq:robust-weight-inv} to the local controller and minimize the difference between the estimated and target end-effector pose, two Cartesian orientation errors are adopted from \cite{schreiber2024crane}. This is done by computing the axis–angle difference between the Z-axes of the current and target end-effector poses. Combined with the Cartesian position errors, the overall pose error is expressed as follows:
\begin{equation}
\label{eq:def-pose-err}
\boldsymbol{e}=\left[\begin{array}{c}
\boldsymbol{e_p} \\
\boldsymbol{e_o}
\end{array}\right]=\left[\begin{array}{lll}
\boldsymbol{p}_{\text {tar }}-
\boldsymbol{p}_{\text {cur }} \\
\cos ^{-1}\left(\frac{\tilde{z}^{\top} z}{\|\tilde{z}\|_2\|z\|_2}\right) \dfrac{\tilde{z}\times z}{\lVert\tilde{z}\times z\rVert_2}
\end{array}\right]
\end{equation}
where $\boldsymbol{e} \in \mathbb{R}^{6 \times 1}$, $\boldsymbol{e_p}, \boldsymbol{e_o} \in \mathbb{R}^{3 \times 1}$. $\boldsymbol{p}_{\text {tar }}$, $\boldsymbol{p}_{\text {cur }} \in \mathbb{R}^{3 \times 1}$ are target and current positions. $\tilde{z}=\boldsymbol{R}_{\text {tar, }[z]}$ and $\boldsymbol{z}=\boldsymbol{R}_{\text {cur, }[z]}$ are the Z-axis vectors of the rotation matrix of target and current pose.
Note that $\boldsymbol{e_o}$ is orthogonal to the needle axis $z$, so its component about that axis vanishes and the task is therefore represented in 6-D but is effectively rank $5$, with rotation about the needle axis left uncontrolled.

Combining \eqref{eq:robust-weight-inv} and \eqref{eq:def-pose-err}, we update the desired joint for local controller as follows:
\begin{equation}
\label{eq:control-ik}
\boldsymbol{q}_{\mathrm{des}} \leftarrow {\boldsymbol{q}_{\text{cur}}}+{J}^{\dagger}_{\mathcal{W}} \boldsymbol{K}_{\mathrm{e}} \boldsymbol{e}
\end{equation}
where $\boldsymbol{K}_{\mathrm{e}} $ is a positive definite diagonal gain matrix, current joint configuration $\boldsymbol{q}_{\text{cur}}$ is estimated from 6-DOF robot base and the microcontrollers for 5-DOF end-effector.

\subsection{Achieving Subgoals with Null-space Control under Joint Constraints}

Our platform is redundant ($n=11$ joints) while the task is represented in $6$-D but has only $5$ independently controlled DOFs (three for tip position and two for the needle-axis direction, with rotation about the needle axis left free). Let $J(\boldsymbol{q})\in\mathbb{R}^{6\times n}$ be the task Jacobian and $J_{\mathcal{W}}^{\dagger}\in\mathbb{R}^{n\times 6}$ the robust weighted pseudo-inverse in \eqref{eq:robust-weight-inv-def}.
The null-space projector is
\begin{equation}
\label{eq:null-projector}
N = I_n - J_{\mathcal{W}}^{\dagger} J \in \mathbb{R}^{n \times n}.
\end{equation}
Strictly, $J(N\boldsymbol{v})=\mathbf{0}$ holds exactly for the undamped projector ($\lambda\to 0$) for any $\boldsymbol{v}\in\mathbb{R}^n$; for the small damping $\lambda$ used here, $JN=\mathcal{O}(\lambda^2)$ and the residual coupling into the primary task is negligible. 
Assuming the task Jacobian attains its structural rank away from singularities ($\mathrm{rank}(J)=5$, since rotation about the needle axis is uncontrolled), the null-space has dimension $n-\mathrm{rank}(J)=6$.

To improve dexterity without affecting the primary task, we maximize Yoshikawa manipulability
\begin{equation}
\label{eq:def-manipu}
w(\boldsymbol{q})=\sqrt{\operatorname{det}\left(J J^T\right)},
\end{equation}
using a gradient-based null-space velocity $\dot{\boldsymbol{q}}_{\text{n}}=\nabla_{\boldsymbol{q}} w(\boldsymbol{q})$. The unconstrained redundancy-resolution command is
\begin{equation}
\label{eq:unconstrained-null}
\dot{\boldsymbol{q}}_{0}
= {J}^{\dagger}_{\mathcal{W}} \boldsymbol{K}_{\mathrm{e}} \boldsymbol{e}
+ N \boldsymbol{K}_{\text{n}} \dot{\boldsymbol{q}}_{\text {n }},
\end{equation}
with positive definite diagonal gains $\boldsymbol{K}_{\mathrm{e}}$ and $\boldsymbol{K}_{\text{n}}$.

Since null-space projection alone does not guarantee joint limits, we enforce joint position and velocity bounds by projecting $\dot{\boldsymbol{q}}_0$ onto the feasible set using a quadratic program at each control cycle:
\begin{equation}
\begin{aligned}
\label{eq:qp_projection}
& \dot{\boldsymbol{q}}^{\star} = \arg\min_{\dot{\boldsymbol{q}}\in\mathbb{R}^n}
\frac{1}{2}\left\|\dot{\boldsymbol{q}}-\dot{\boldsymbol{q}}_0\right\|_2^2 \\
\text{s.t.} \quad 
& \dot{\boldsymbol{q}}_{\min} \le \dot{\boldsymbol{q}} \le \dot{\boldsymbol{q}}_{\max},\\
& \boldsymbol{q}_{\min} \le \boldsymbol{q}_{\text{cur}} + \Delta t \dot{\boldsymbol{q}} \le \boldsymbol{q}_{\max},
\end{aligned}
\end{equation}
where $\Delta t$ is the control timestep. This yields hard enforcement of joint motion limitations while remaining close to the desired null-space augmented command.

Finally, the joint update rule becomes
\begin{equation}
\label{eq:control-ik+null}
\boldsymbol{q}_{\mathrm{des}} \leftarrow {\boldsymbol{q}_{\text{cur}}}+ \Delta t \, \dot{\boldsymbol{q}}^{\star}.
\end{equation}

\section{EXPERIMENTS}


\subsection{In-bore Reachability in Simulated Clinical Scenarios}


\begin{figure}[t]
    \centering
    \begin{subfigure}[t]{0.499\textwidth}
        \centering
        \includegraphics[width=0.8\textwidth]{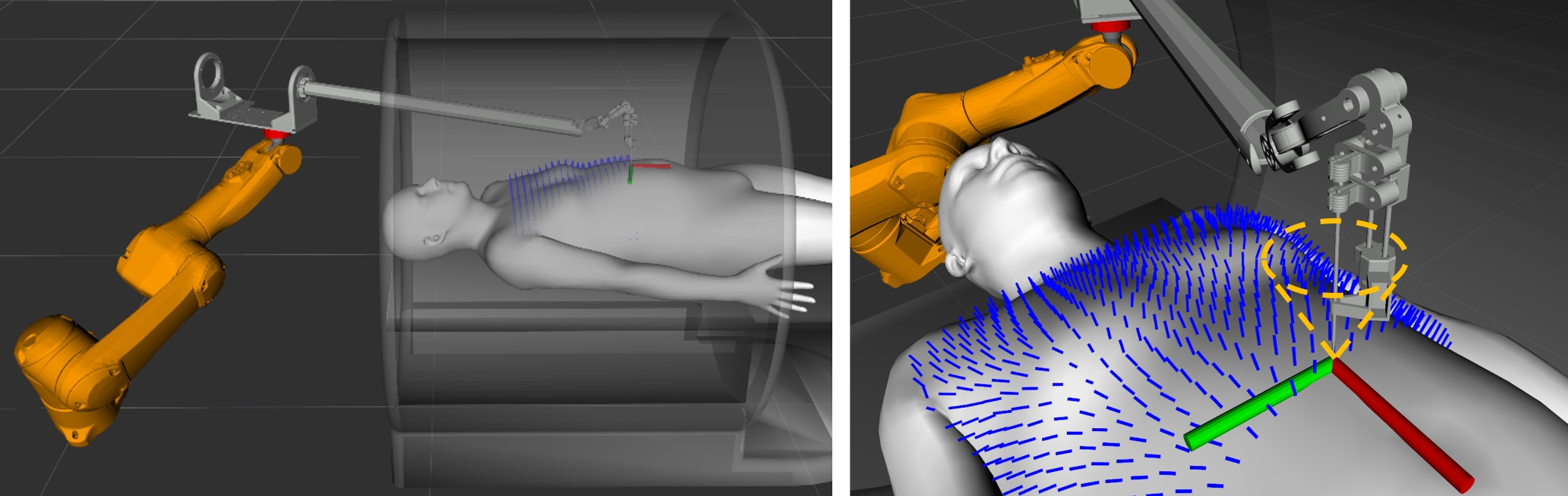}
        \caption{Simulation setup for reachability evaluation}
        \label{fig:reach-subfig-scene}
    \end{subfigure}
    \begin{subfigure}[t]{0.499\textwidth}
        \centering
        \includegraphics[width=0.99\textwidth]{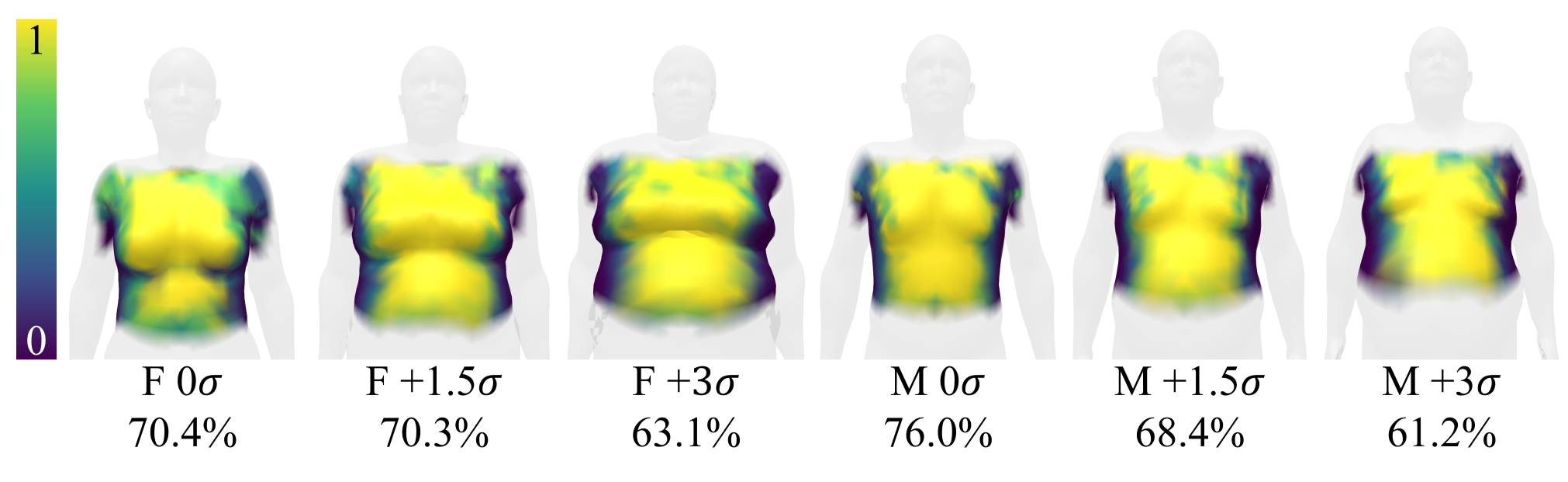}
        \caption{Reachable result of previous 8-DOF system}
        \label{fig:reach-subfig-old}
    \end{subfigure}
    \begin{subfigure}[t]{0.499\textwidth}
        \centering
        \includegraphics[width=0.99\textwidth]{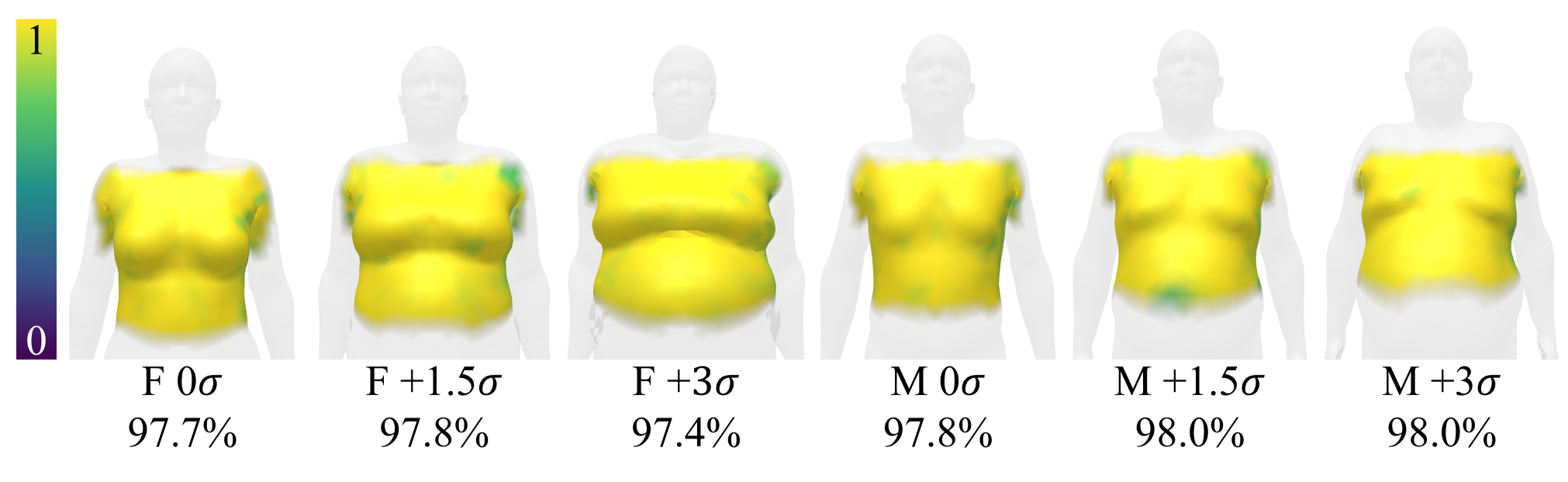}
        \caption{Reachable result of new 11-DOF system}
        \label{fig:reach-subfig-new}
    \end{subfigure}
    \begin{subfigure}[t]{0.499\textwidth}
        \centering
        \includegraphics[width=0.99\textwidth]{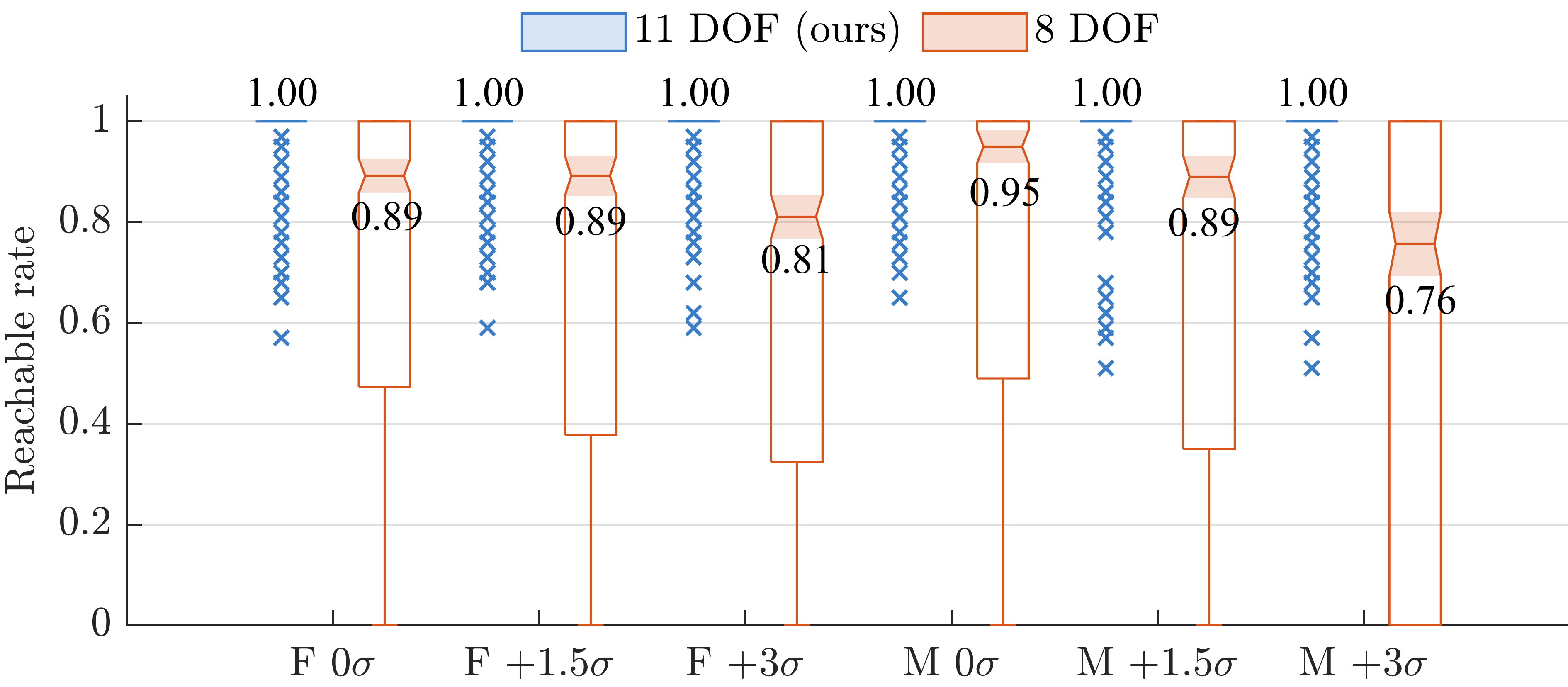}
        \caption{Reachable rate across samples over patient chest}
        \label{fig:reach-subfig-box}
    \end{subfigure}
    \caption{
    {Comparison of in-bore reachability between previous 8-DOF and new 11-DOF system. The color brightness on mesh surface shows reachable rate at each point. F is female and M is male, respectively. Larger patients result in less in-bore space and a more challenging environment.}
    }
    \label{fig:cmp-reach}
\end{figure}

To quantitatively evaluate the dexterity advantages of the new 11-DOF system, we conducted a comparative analysis of in-bore reachability across multiple generated human anatomies against the previous 8-DOF design. The experiments assessed practical workspace coverage in clinically relevant scenarios.
The MoveIt IK solver was used to obtain collision-free joint configurations.

First, human bodies of different sizes were generated using STAR \cite{osman2020star}, including average (+0$\sigma$), large (+1.5$\sigma$), and very large (+3$\sigma$) sizes for both females and males.
These bodies were then placed within the scanner bore in a position accessible to the needle-biopsy robot.
Mesh normals were generated over the chest area to represent candidate biopsy targets, and additional nearby orientations around each nominal normal were added as shown in Fig.~\ref{fig:reach-subfig-scene}.
An inverse kinematics solver was used to compute reachability for all targets. At each normal, the reachability rate was computed and mapped to color at the corresponding mesh location for visualization.
Specifically, the scanner bore diameter was 85 cm; 600 normals were used, and 36 additional orientations within a 30$^\circ$ adjustment were added at each normal. The reachability error tolerance were 1 mm for position and 1$^\circ$ for orientation.

Figures ~\ref{fig:reach-subfig-old} and \ref{fig:reach-subfig-new} illustrate the robot’s ability to dexterously insert needles across diverse patient anatomies and angles is shown, with brighter colors indicating higher reachability.
The new 11-DOF system maintained a consistently high reachability rate, exceeding 97\% across all scenarios.
In contrast, the 8-DOF system exhibited obvious non-reachable regions concentrated along both sides of the torso, where proximity to the scanner bore imposed greater constraints.
The reachability of the 8-DOF system also degraded significantly with larger anatomies due to reduced in-bore space and a more constrained environment.
This trend is also evident in the box chart in Fig.~\ref{fig:reach-subfig-box}. Note that the 11-DOF system achieved full reachability across the area of interest in all scenarios except a few outliers, whereas the 8-DOF system exhibited considerable number of low-reachable and non-reachable cases.
These findings highlight limitations of the prior 8-DOF design in accommodating patient variability and confirm that the enhanced dexterity and expanded workspace of the new system translate to improved reachability in realistic clinical settings.

\subsection{Task-oriented Weight Policy Investigation}
\label{sec:exp-weight}
\begin{figure*}[t!]
    \centering
    \begin{subfigure}[t]{0.195\textwidth}
        \centering
        \includegraphics[width=0.9\textwidth, height=0.645\textwidth]{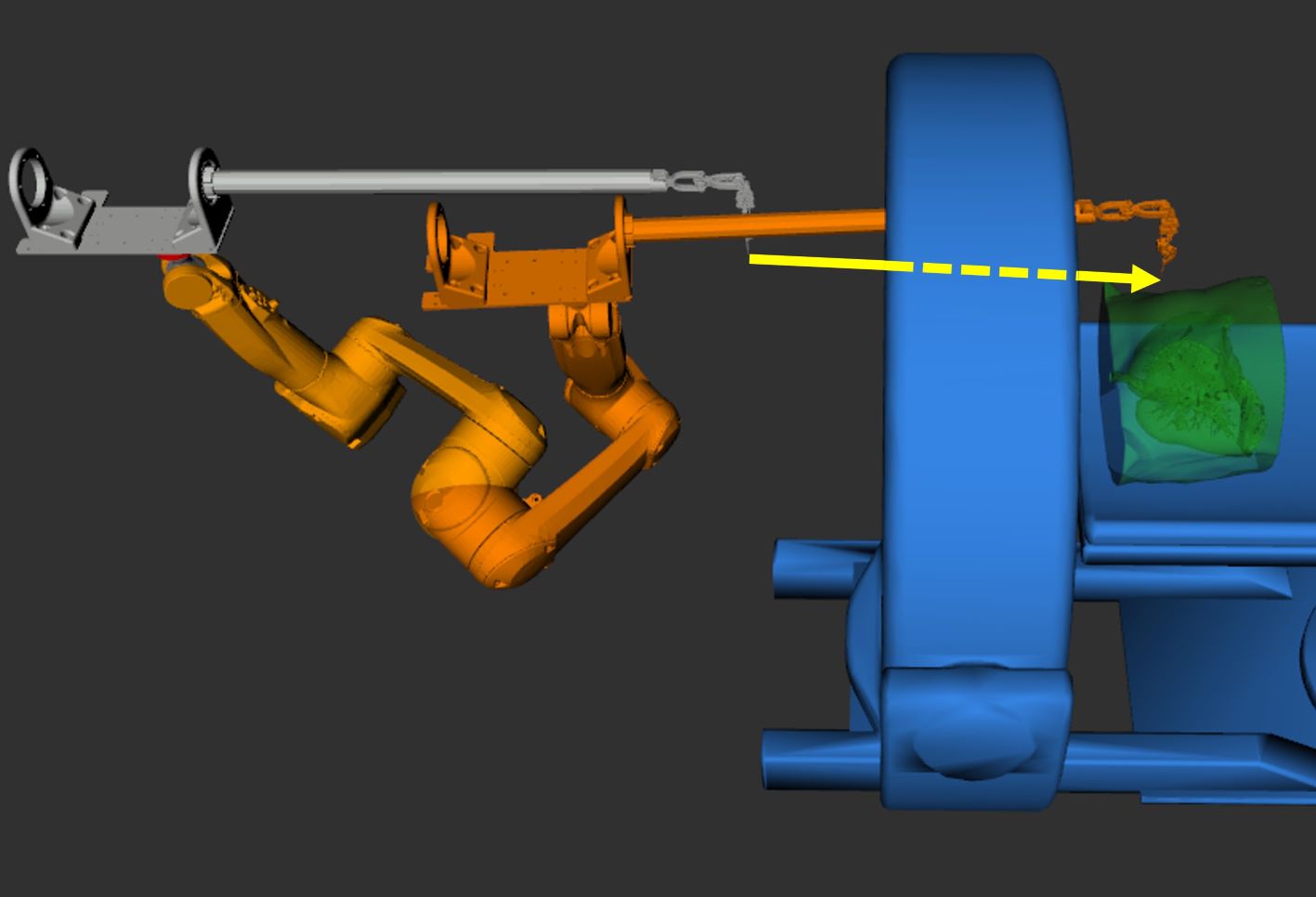}
        \vspace{2pt} 
        \includegraphics[width=\textwidth]{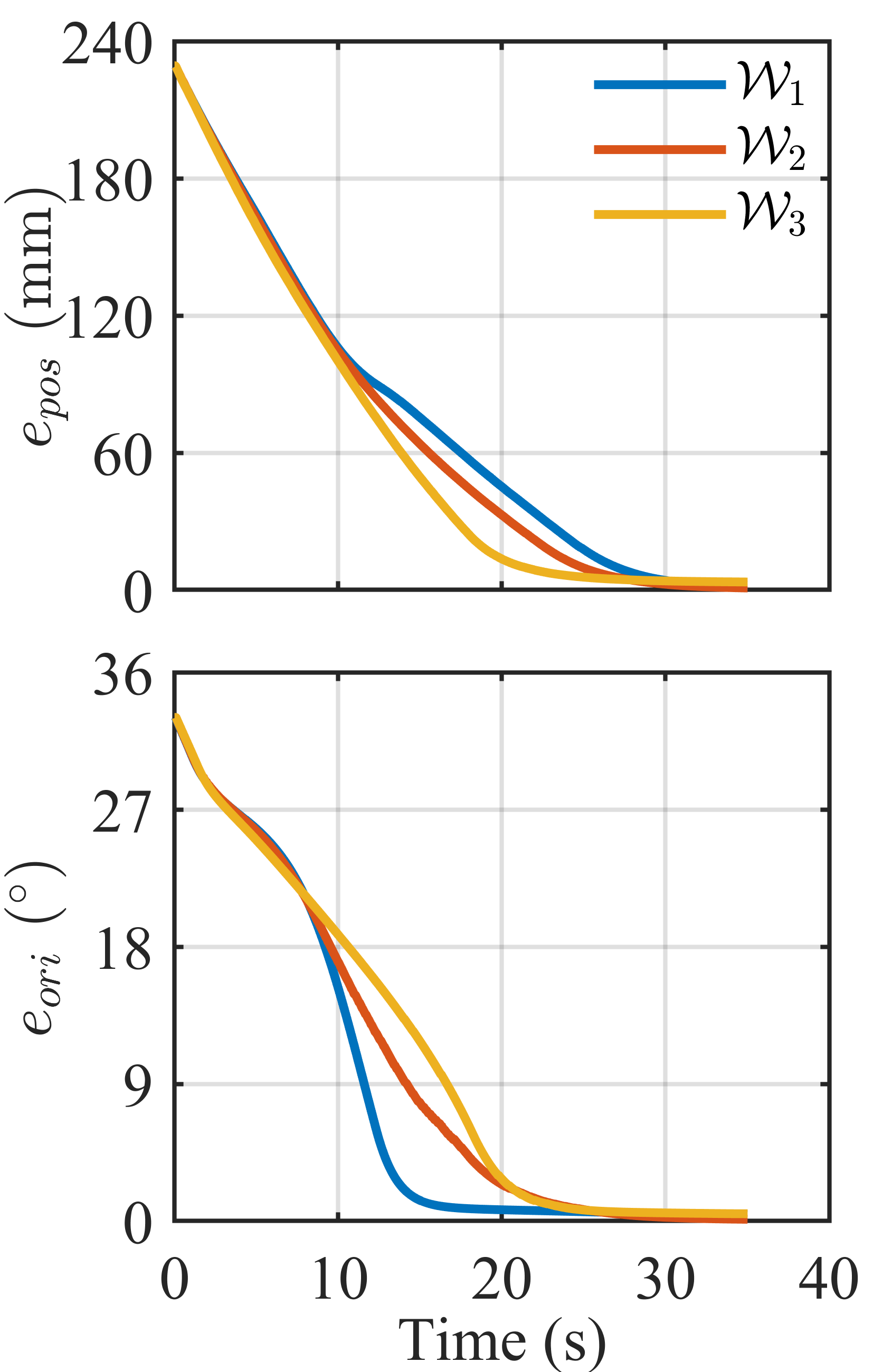} 
        \caption{Reaching In-Bore}
        \label{fig:subfigA}
    \end{subfigure}
    \hfill
    \begin{subfigure}[t]{0.195\textwidth}
        \centering
        \includegraphics[width=0.9\textwidth]{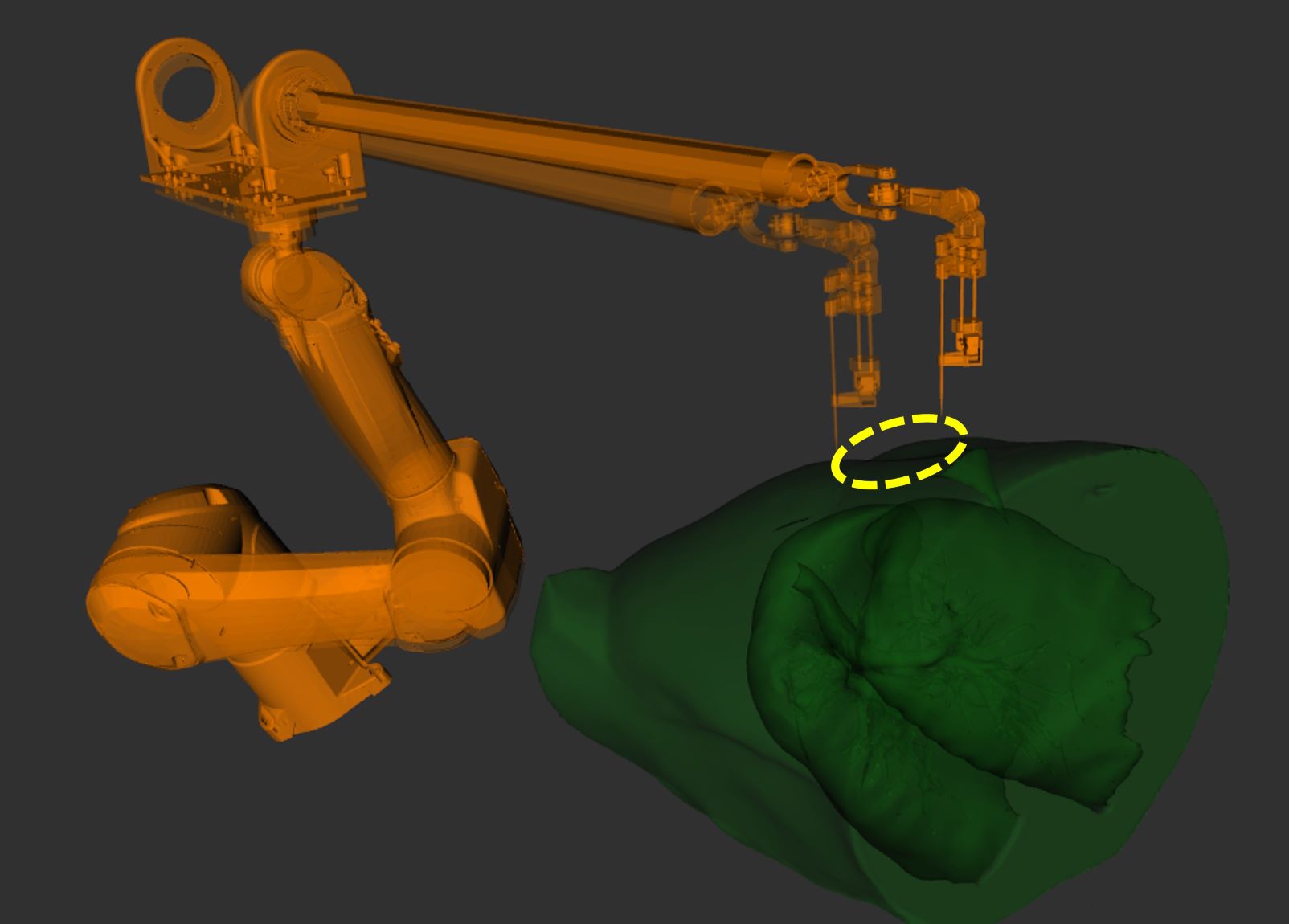}
        \vspace{2pt} 
        \includegraphics[width=\textwidth]{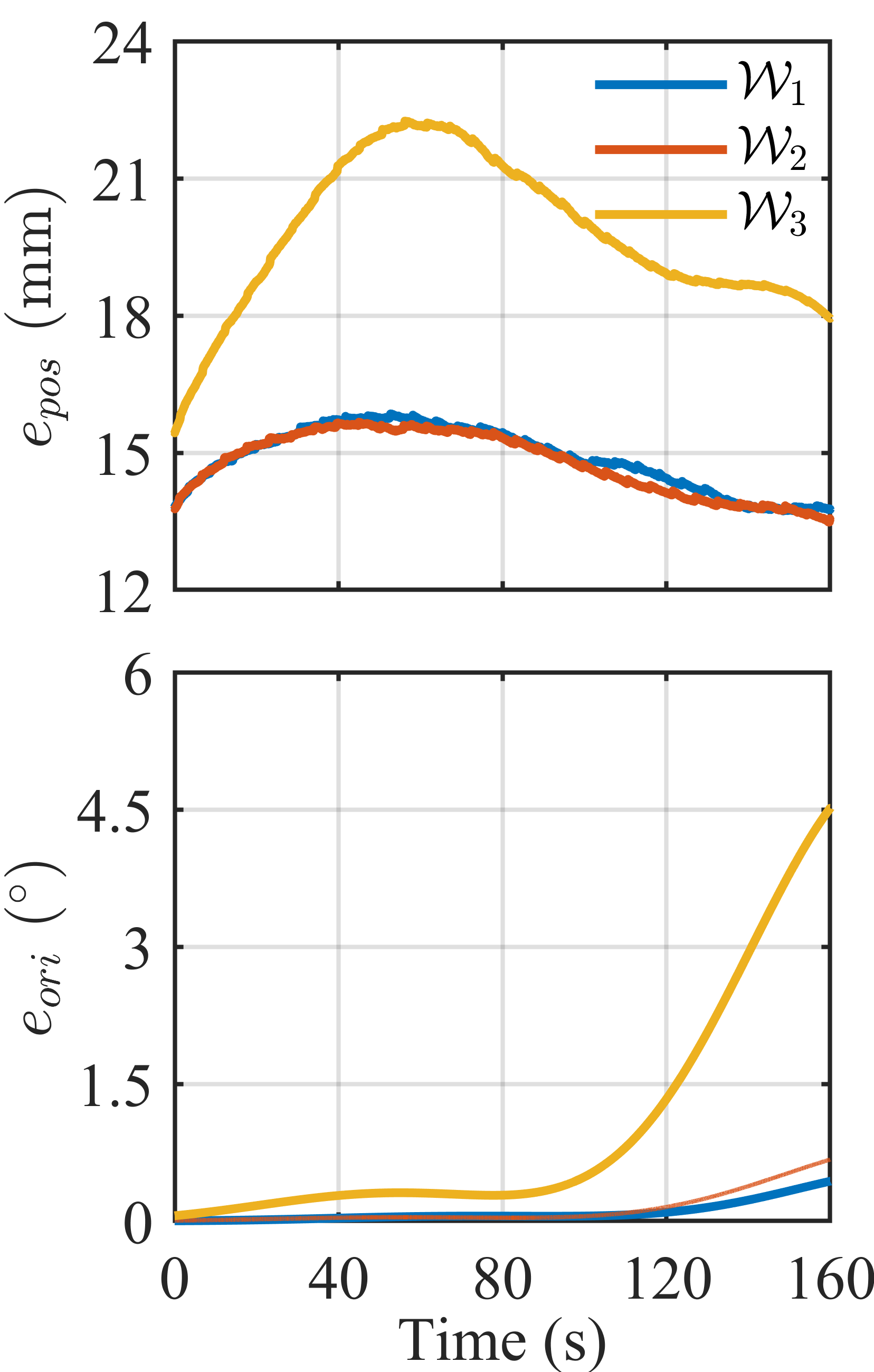} 
        \caption{In-Bore Positioning}
        \label{fig:subfigB}
    \end{subfigure}
    \begin{subfigure}[t]{0.195\textwidth}
        \centering
        \includegraphics[width=0.9\textwidth]{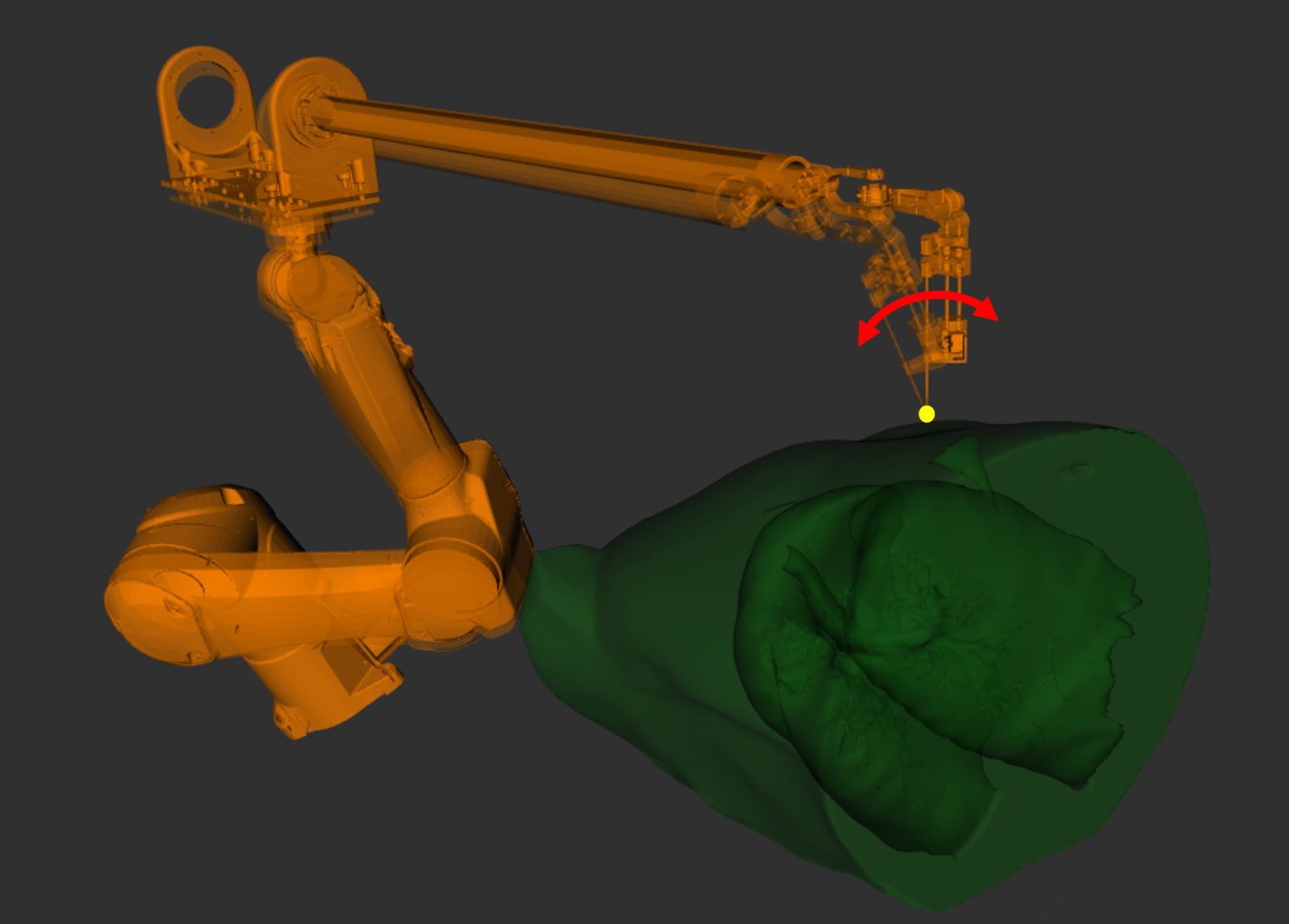}
        \vspace{2pt} 
        \includegraphics[width=\textwidth]{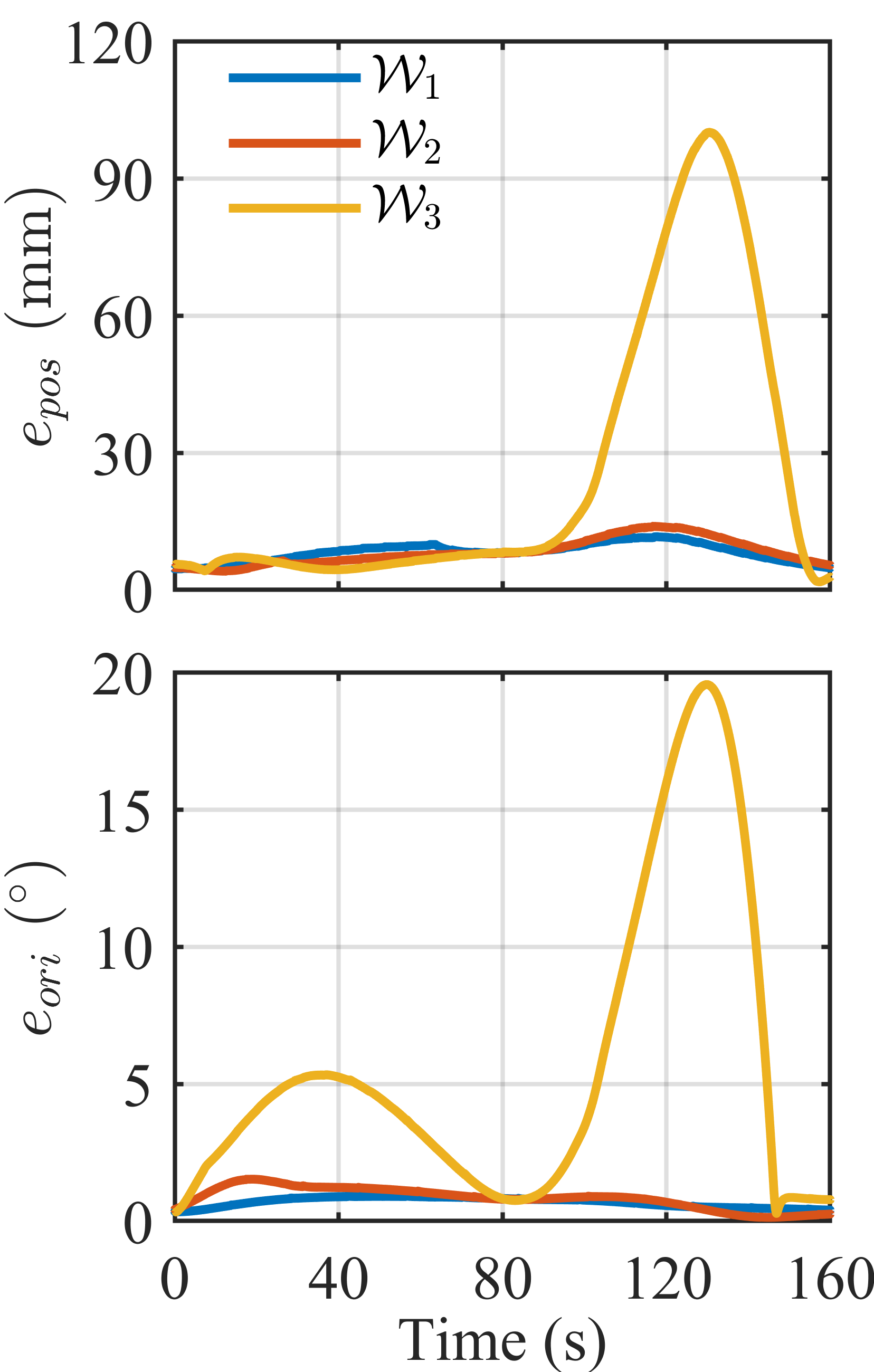}
        \caption{In-Bore RCM}
        \label{fig:subfigC}
    \end{subfigure}
    \hfill
    \begin{subfigure}[t]{0.195\textwidth}
        \centering
        \includegraphics[width=0.9\textwidth]{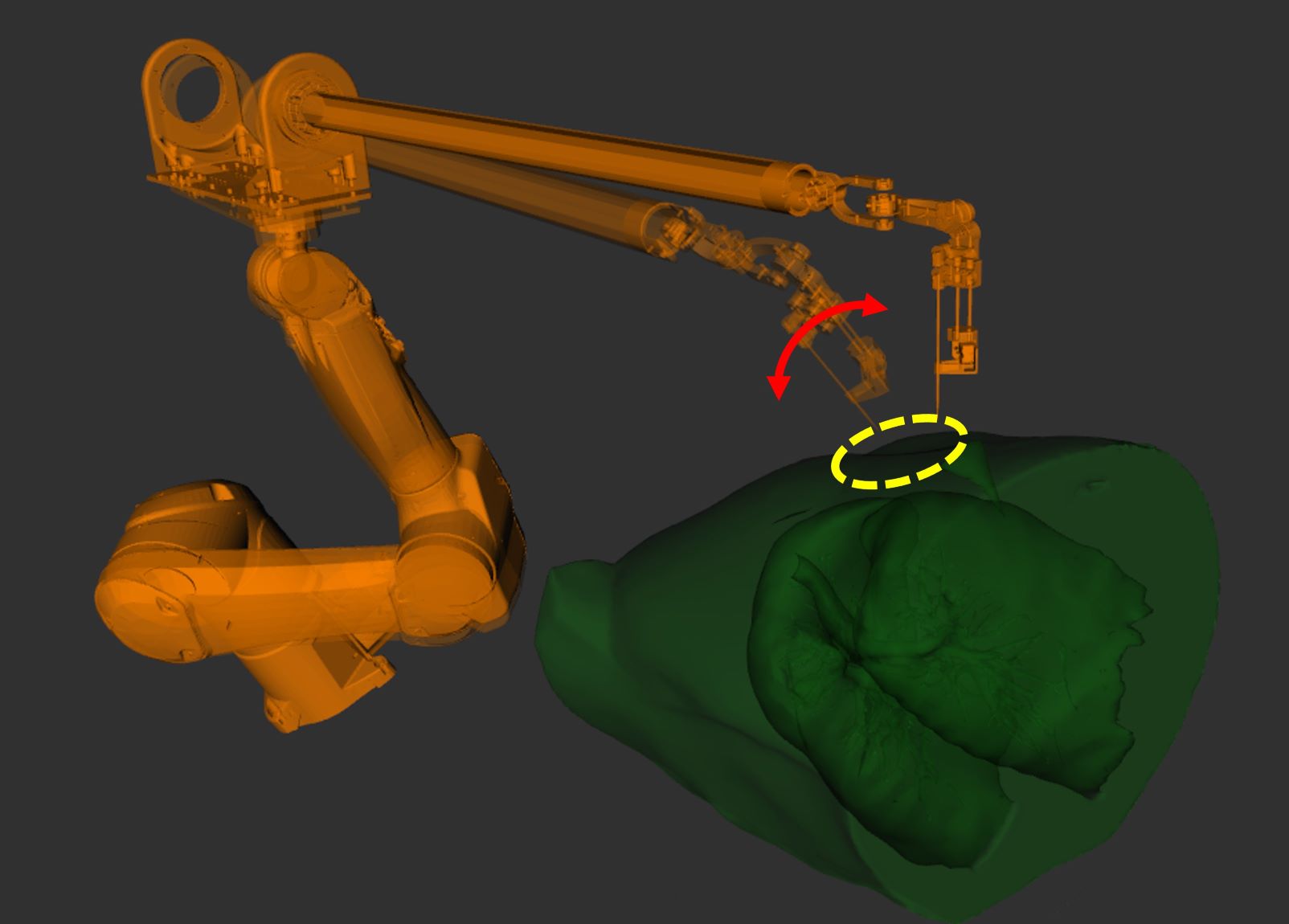}
        \vspace{2pt} 
        \includegraphics[width=\textwidth]{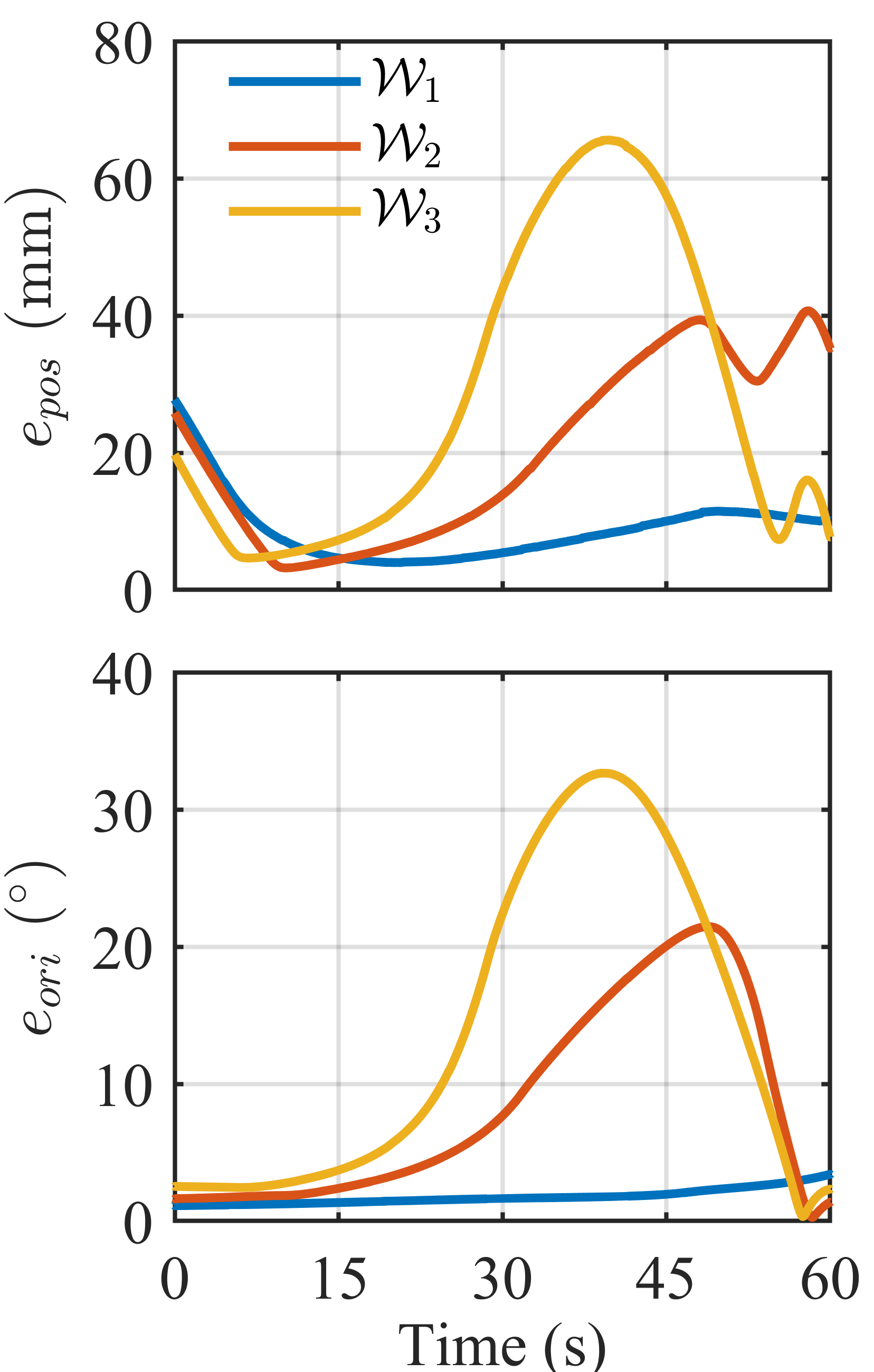}
        \caption{In-Bore Pose Tracking}
        \label{fig:subfigD}
    \end{subfigure}
    \hfill
    \begin{subfigure}[t]{0.195\textwidth}
        \centering
        \includegraphics[width=0.9\textwidth]{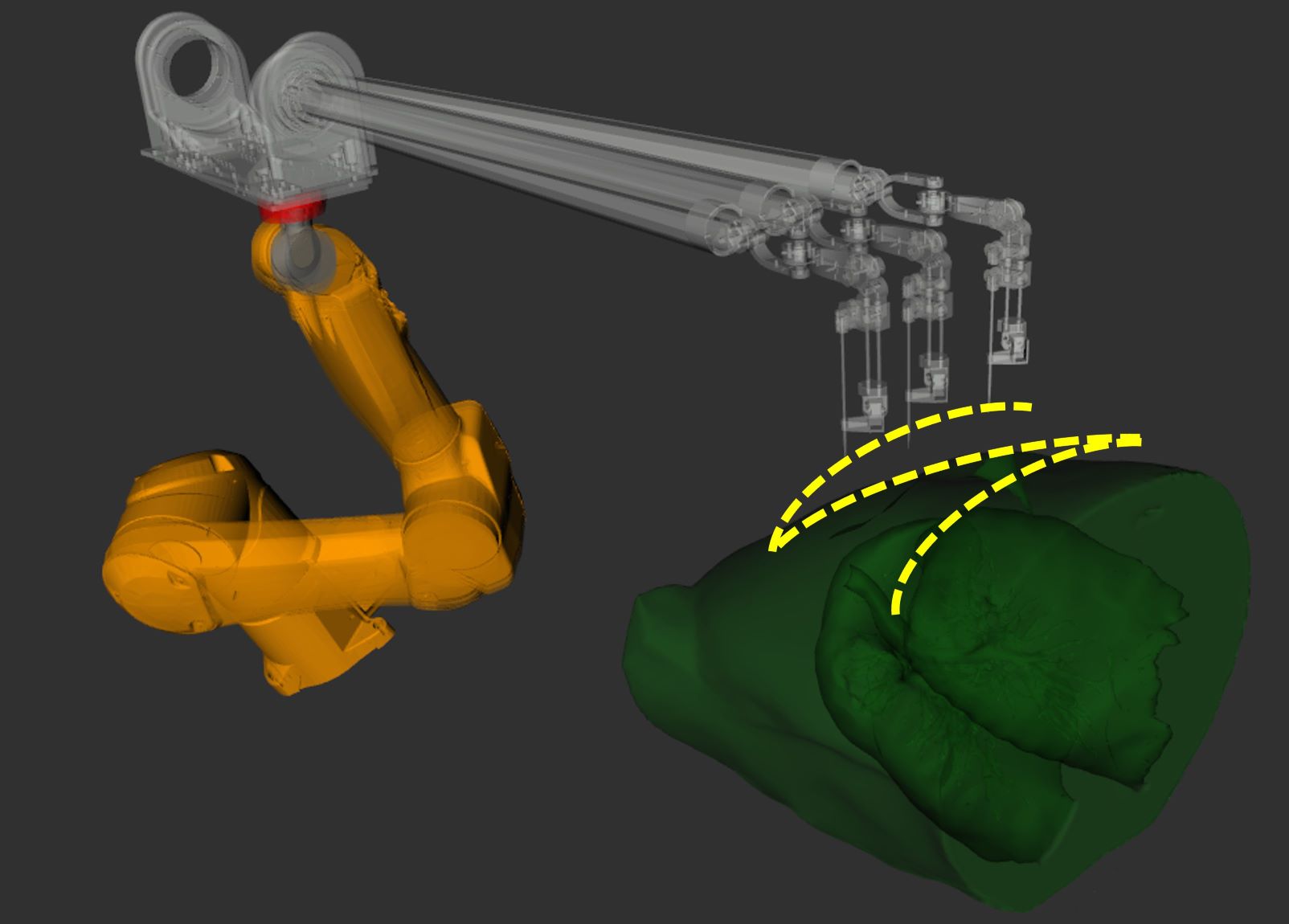}
        \vspace{2pt} 
        \includegraphics[width=\textwidth]{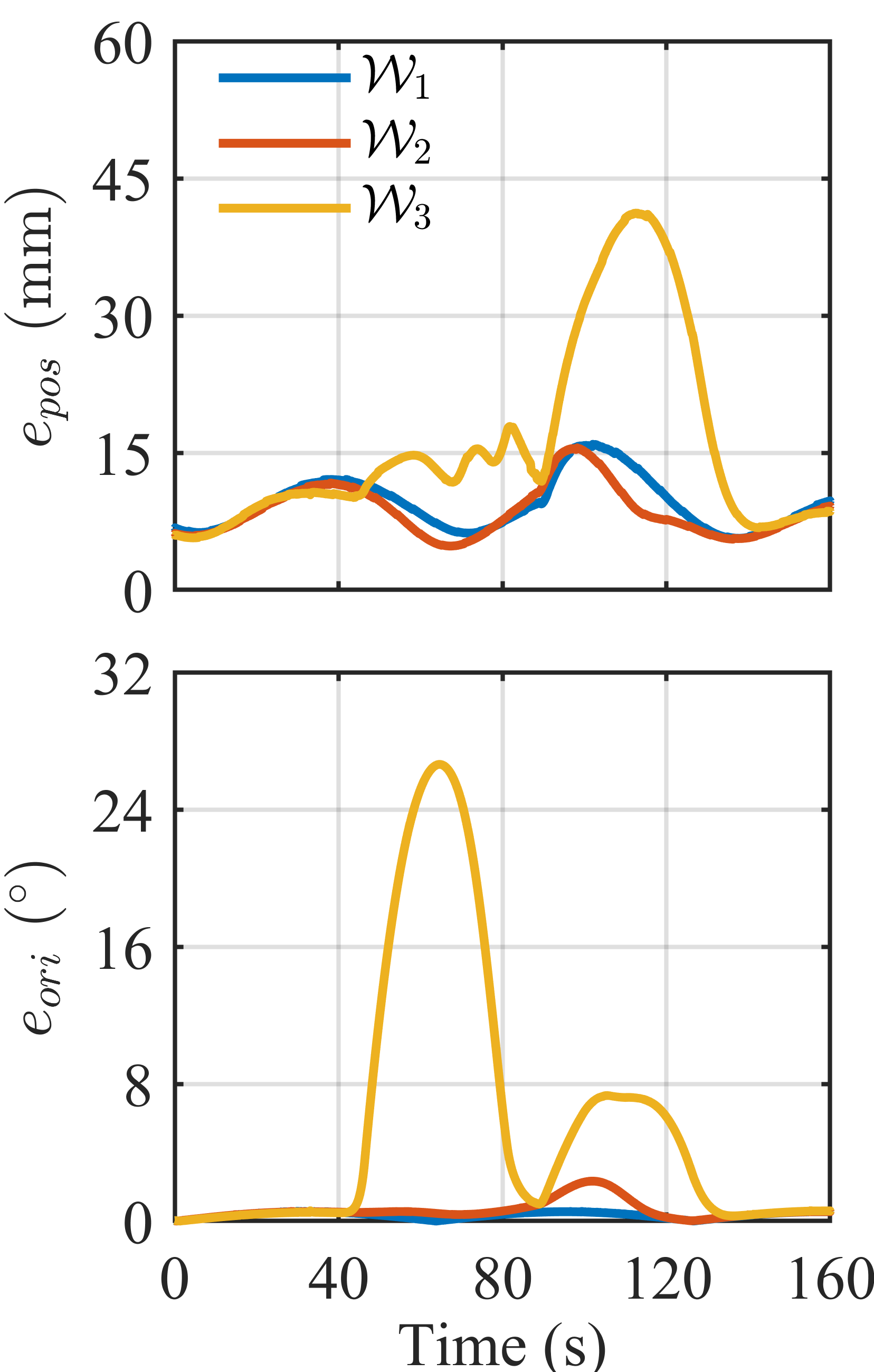} 
        \caption{In-Bore Z-Trajectory}
        \label{fig:subfigE}
    \end{subfigure}
    \caption{Comparison of tracking error of 5 different trajectories with 3 weight policies.
    While the experiments were conducted in the real world, the top row presents simulation figures to illustrate the trajectory in the task.
    For the largest movement, reaching into the bore, $\mathcal{W}_3$ is able to translate the fastest since the 6-DOF robotic base is used.
    Meanwhile for in-bore manipulation, $\mathcal{W}_1$ is able to converge faster since the 5-DOF end-effector is used for the fine precision movements.
    }
    \label{fig:cmp-track-perf}
\end{figure*}

We validate the weighted inverse-Jacobian control by examining five representative trajectories in clinical biopsy scenarios.
The weight matrix $\mathcal{W}$ in \eqref{eq:robust-weight-inv-def} is treated as a hyperparameter, and its effects on biopsy performance are evaluated.
Due to the modular design of the 11-DOF system, the weight matrix is divided into two parts for study: the 6-DOF base and the 5-DOF end-effector. An immediate baseline uses equal weights represented by the identity matrix. The other two cases emphasize either the base or the end-effector. To demonstrate these effects, a factor of 10 difference in weights is applied. As a result, three weight policies are used:
$\mathcal{W}_1= [\mathbf{I}^{6 \times 6}, \mathbf{0}^{6 \times 5}; \mathbf{0}^{5 \times 6}, 0.1 \times \mathbf{I}^{5 \times 5}], ~
\mathcal{W}_2= \mathbf{I}^{11 \times 11}, ~
\mathcal{W}_3= [0.1 \times \mathbf{I}^{6 \times 6}, \mathbf{0}^{6 \times 5}; \mathbf{0}^{5 \times 6},  \mathbf{I}^{5 \times 5}] \in \mathbb{R}^{11 \times 11}$ where $\mathbf{I}$ is identity matrix and $ \mathbf{0}$ is zero matrix.
 $\mathcal{W}_1$ emphasizes end-effector motion by assigning lower weights to the end effector, $\mathcal{W}_2$ corresponds to unweighted control, and $\mathcal{W}_3$ emphasizes base motion.
Other parameters proven robust in both simulation and real-world tests are: control gains $\boldsymbol{K}_{\mathrm{e}} = 0.02,\mathbf{I}^{6 \times 6}$, $\boldsymbol{K}_{\mathrm{n}} = 0.02,\mathbf{I}^{11 \times 11}$, and damping factor $\lambda = 10^{-4}$.

The tracking performance is characterized by the mean position and orientation errors, as reported in Fig. \ref{fig:cmp-track-perf} and Table \ref{tab:cmp}.
$T_{r,\mathrm{pos}}$ and $T_{s,\mathrm{pos}}$ denote the rise time and settling time of position tracking. $e_{ss}$ is the steady-state error; $e_{\mathrm{pos}}$ and $e_{\mathrm{ori}}$ are the position and orientation errors.
During the experiments, the robotic platform is controlled to track five typical trajectory scenarios, as shown in the first row of Fig.~\ref{fig:cmp-track-perf}.
RRT-Connect is used to generate collision-free path for the reach-in-bore stage. 
The trajectories include:
\begin{enumerate}
\item \textbf{Reaching In-Bore}: Step response for a fixed pose;
\item \textbf{In-Bore Positioning}: Circular motion in position with a radius of 5 cm;
\item \textbf{In-Bore RCM}: Orientation tracking with a maximum angular deviation of 45$^\circ$ from the center;
\item \textbf{In-Bore Pose Tracking}: Combined position and orientation motion, involving a 5 cm-radius circular trajectory and a maximum angular deviation of 45$^\circ$ from the center;
\item \textbf{In-Bore Z-Trajectory}: A ``z-shape'' path (position and orientation) covering a 20 cm $\times$ 20 cm area in the $xy$-plane, a maximum height difference of 5 cm in $z$, and a maximum angular deviation of 45$^\circ$.
\end{enumerate}
These trajectories can be categorized into two primary stages during a clinical CT-guided needle insertion procedure.
Reaching In-Bore corresponds to the initial large-scale gross motion of the robot platform from outside the scanner to inside the imaging bore to approach the patient.
This stage faces fewer constraints and requires a faster response to minimize overall operation time. The other trajectories represent the in-bore fine manipulation task and adjustment of the needle, which demands accurate and careful movement within the narrow imaging bore and the patient’s chest.
By comparing the real-time robot end-effector trajectories obtained via forward kinematics with the desired trajectories, we observe different performances for the three weight policies across these five scenarios.

\begin{table}[t]
\centering
\caption{Comparison of trajectory tracking performance under three different weight policies}
\label{tab:cmp}
\begin{tblr}{
  width = \linewidth,
  colspec = {Q[279]Q[231]Q[135]Q[135]Q[135]},
  cells = {c},
  cell{1}{1} = {r=2}{},
  cell{1}{2} = {r=2}{},
  cell{1}{3} = {c=3}{0.405\linewidth},
  cell{3}{1} = {r=4}{},
  cell{7}{1} = {r=2}{},
  cell{9}{1} = {r=2}{},
  cell{11}{1} = {r=2}{},
  cell{13}{1} = {r=2}{},
  vline{2-3} = {1-14}{},
  vline{3} = {4-6,8,10,12}{},
  hline{1,3,7,9,11,13,15} = {-}{},
}
Trajectory             & Metric             & Weight Policy &        &        \\
                       &                    & $\mathcal{W}_1$            & $\mathcal{W}_2$     & $\mathcal{W}_3$     \\
{Reaching In-Bore}        & $T_{r, pos}$ (s)           & 22.3         & 20.2  & \textbf{16.6}  \\
                       & $T_{s, pos}$ (s)           & 26.5         & 24.4  & \textbf{20.7}  \\
                       & $e_{ss, pos}$ (mm)          & 2.0        & \textbf{0.8} & 3.5 \\
                       & $e_{ss, ori}$ ($^{\circ}$)        & 0.4       & \textbf{0.1} & 0.5 \\
{In-Bore\\Positioning}  & $\bar e_{pos}$ (mm)     & 14.9        & \textbf{14.8} & 19.8 \\
                       & $\bar e_{ori}$ ($^{\circ}$) & \textbf{0.1}        & 0.1 & 1.0 \\
{In-Bore \\ RCM} & $\bar e_{pos}$ (mm)     & \textbf{8.3}        & 8.4 & 25.3 \\
                       & $\bar e_{ori}$ ($^{\circ}$) & \textbf{0.7}        & 0.8 & 5.7 \\
{In-Bore Pose \\ Tracking}     & $\bar e_{pos}$ (mm)     & \textbf{8.8}        & 20.1 & 28.2 \\
                       & $\bar e_{ori}$ ($^{\circ}$) & \textbf{1.8}        & 8.5 & 13.8 \\
{In-Bore \\ Z-Trajectory}     & $\bar e_{pos}$ (mm)     & {9.4}        & \textbf{8.6} & 16.1 \\
                   & $\bar e_{ori}$ ($^{\circ}$) & \textbf{0.3}        & 0.6 & 5.8
\end{tblr}
\end{table}

\subsubsection{Reaching In-Bore}
$\mathcal{W}_3$ achieves the shortest rise time and settling time in position, reducing these times by up to $25.6 \%$ and $17.8 \%$ compared with $\mathcal{W}_1$ and $\mathcal{W}_2$, respectively. This is due to the lower penalty weight on the robot base in $\mathcal{W}_3$, which facilitates large-scale end-effector motion. Despite this faster response, $\mathcal{W}_3$ maintains acceptable steady state position and orientation errors, suggesting that rapid gross movement can be safely applied when approaching the patient inside the imaging bore. Conversely, $\mathcal{W}_1$ achieves the shortest rise and settling times in orientation but the longest in position. $\mathcal{W}_2$ exhibits the least steady-state position and orientation errors overall.

\subsubsection{In-Bore Manipulation}
$\mathcal{W}_1$ proves more favorable by providing the highest orientation accuracy in all trajectories and the highest or second-highest position accuracy. Because the robot base handles $xyz$-axis motions at this stage and the end-effector tracks orientation, $\mathcal{W}_1$ 's lower penalty on end-effector joints allows it to respond more quickly to control commands and precisely track the desired orientation. In contrast, $\mathcal{W}_3$ shows the lowest accuracy during fine in-bore motion. Its lower penalty on robot base movement leads to overshoot in small-scale movements, making it nearly unusable for this stage due to its large errors. Additionally, as seen in Fig. \ref{fig:cmp-track-perf}, $\mathcal{W}_3$ experiences spikes in orientation and position errors during prolonged orientation tracking.
In our experiments, these events coincided with proximity to joint limits, suggesting a transient degradation in kinematic conditioning, likely because $\mathcal{W}_3$ tends to compensate more with robot base movement.

In summary, a low penalty weight on the robot base (like $\mathcal{W}_3$) leads to faster tracking response and is advantageous during the gross-motion phase of needle insertion. Conversely, a low penalty weight on the end-effector (like $\mathcal{W}_1$) achieves better tracking accuracy and is more suitable for fine needle manipulation within the confined imaging bore.
Note that although we identify a qualitative relationship between weight-policy selection and accuracy, the weights used here are not optimized, and finding optimal weights remains an open problem because analytic relations between trajectory-tracking accuracy and weight parameters are difficult to establish. Moreover, optimizing other metrics, such as energy consumption, may favor different parameter choices.



\begin{figure}[t]
\centering
\includegraphics[width=0.99\linewidth]{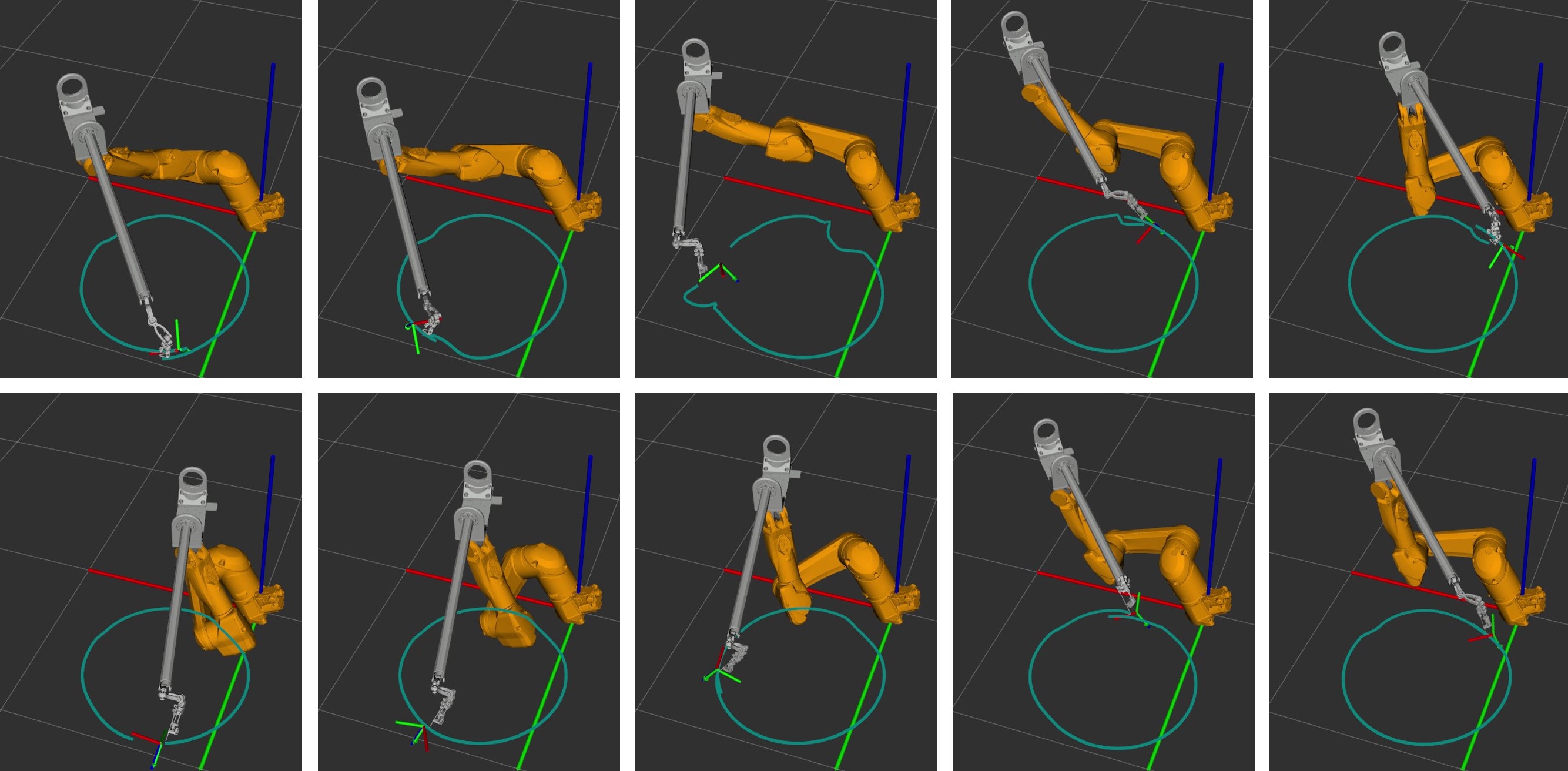}
\caption{Comparison of joint configurations between robots without (first row) and with (second row) null-space control. The end-effector deviates from the desired circle trajectory due to reaching a singularity without null-space control.}
\label{fig:comp-null-space}
\end{figure}
\begin{figure}[t]
    \centering
    \includegraphics[width=0.49\textwidth, height=0.17\textwidth]{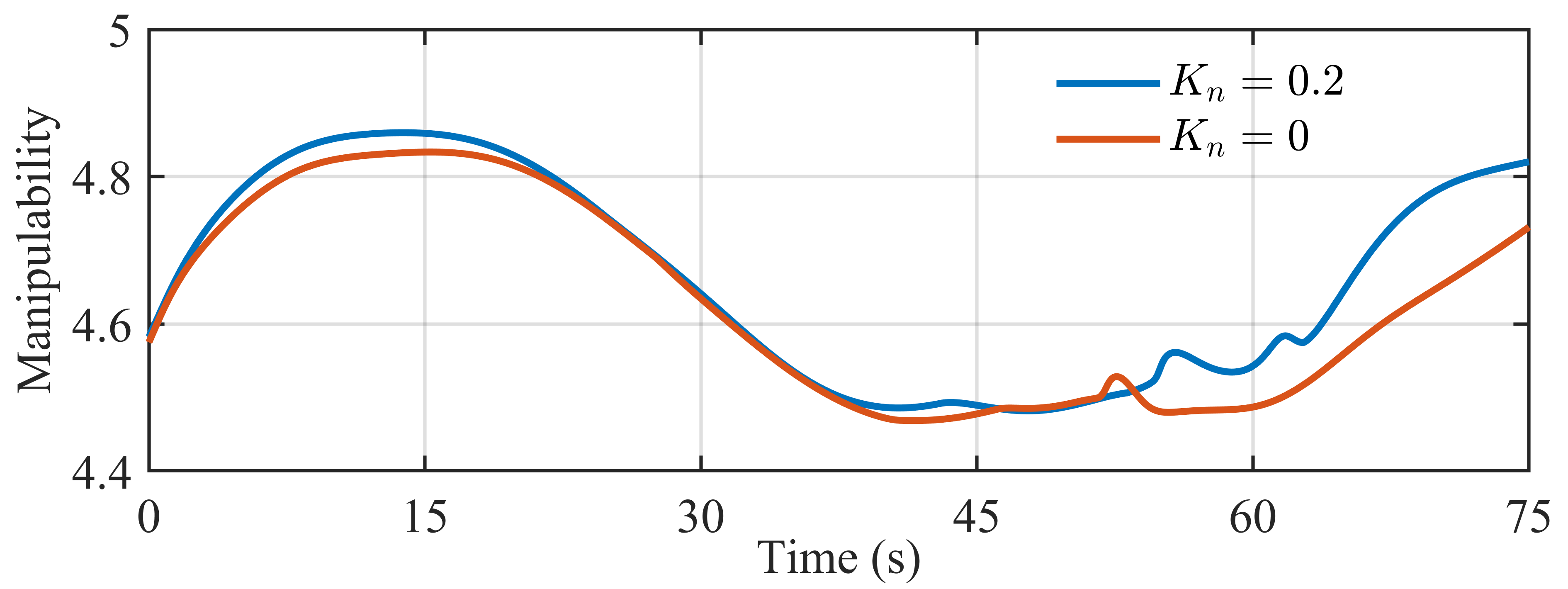}
    \caption{Comparison of Yoshikawa manipulability measure with and without null-space control.}
    \label{fig:null-manipu}
\end{figure}
\begin{figure}[t!]
    \centering
    \includegraphics[width=0.49\textwidth, height=0.17\textwidth]{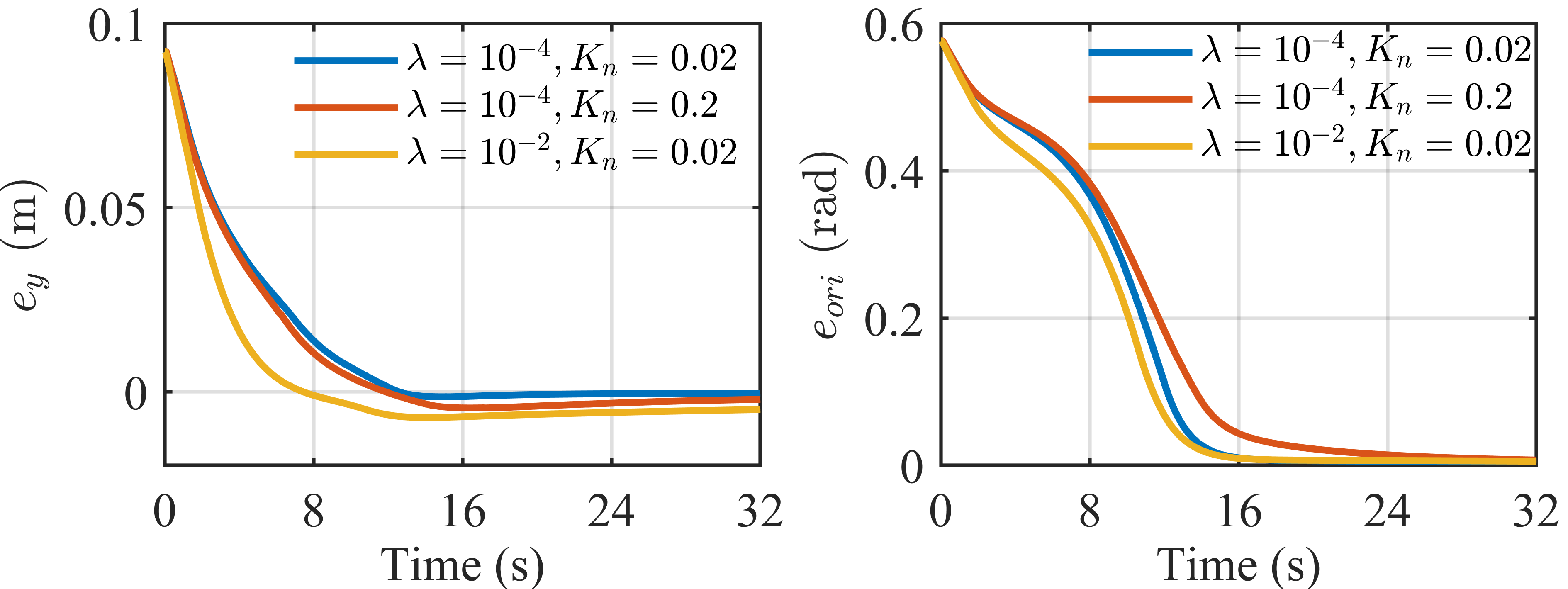}
    \caption{Comparison of step response accuracy under different null-space control gain $\boldsymbol{K}_{\mathrm{n}}$ and damping factor $\lambda$.}
    \label{fig:null-param}
\end{figure}

\begin{figure*}[t!]
    \centering
    \includegraphics[width=0.99\textwidth]{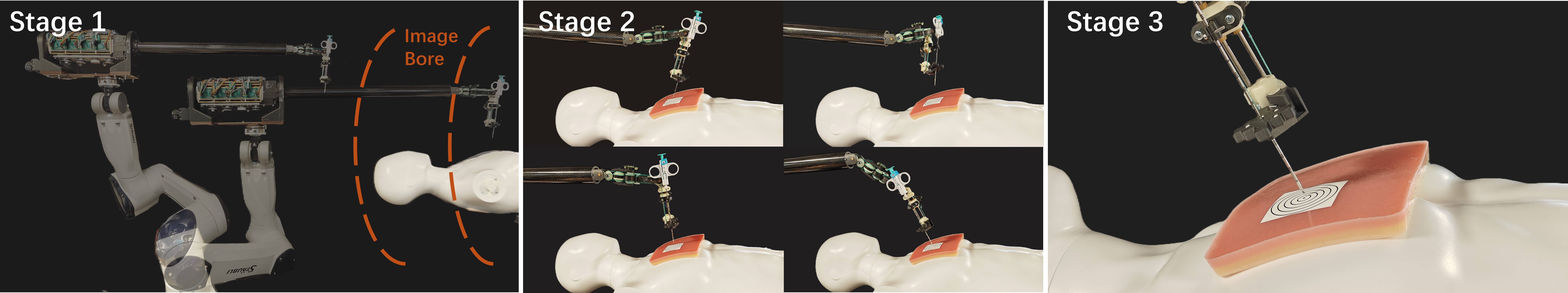}
    \caption{Demonstration of tele-operational control of our dexterous robot platform. The workflow consists of three stages: 1) large-scale motion to reach in-bore; 2) fine in-bore adjustment of needle pose; 3) precise needle insertion.}
    \label{fig:teleop}
\end{figure*}

\subsection{Null-space Controller}

Next, we demonstrate the effect of null-space control by comparing the robot arm configurations with and without null-space control. As shown in Fig.~\ref{fig:comp-null-space}, the end-effector follows a circular trajectory with a 30 cm radius.
Without null-space control, the robot’s joint configuration tends to stretch into a straight-line posture, resulting in reduced manipulability and increased end-effector deviation from the trajectory due to singularities. In contrast, with null-space control, the configuration remains bent, leading to higher manipulability and reduced tracking deviation.
Figure~\ref{fig:null-manipu} presents the Yoshikawa manipulability measure for both methods, demonstrating that the robot with null-space control achieves a higher manipulability value.

We also investigate the impact of null-space control parameters, namely the null-space control gain $\boldsymbol{K}_{\mathrm{n}}$ and damping factor $\lambda$, on tracking accuracy. The step response of a fixed target pose is used as an example, as shown in Fig. \ref{fig:null-param}.
An increase in steady-state position error is observed with larger values of $\lambda$ and $\boldsymbol{K}_{\mathrm{n}}$.
$\lambda$ is introduced for numerical stability in the Levenberg-Marquardt (LM) algorithm. However, an excessively large $\lambda$ can introduce numerical errors in the computation of the projection matrix, leading to a null-space that is not strictly orthogonal. Consequently, the primary task tracking is affected, resulting in a larger tracking error.
Additionally, if the secondary task is assigned an excessively high $\boldsymbol{K}_{\mathrm{n}}$ relative to the primary task gain $\boldsymbol{K}_{\mathrm{e}}$, the manipulator may move aggressively in the null-space, thereby degrading the tracking performance of the primary task.
Thus, selecting an appropriate damping factor requires balancing numerical stability of computing pseudo-inverse and the numerical error in computing the projection matrix. Moreover, a properly tuned null-space control gain is essential to ensure that the primary task remains unaffected.

\subsection{Teleoperation Demonstration}
Finally, we validate the robot platform and control algorithm through a teleoperation test. In this experiment, the operator uses a Phantom Omni device to provide the target pose for the robot end-effector's needle tip. The workflow, illustrated in Fig.~\ref{fig:teleop}, consists of three stages: (1) maneuvering the end-effector from the setup position outside the bore to inside the bore; (2) adjusting the needle’s position and orientation within the imaging bore; and (3) inserting the needle into the target location.
This test demonstrates the capability of our robotic platform to accurately achieve target positions and successfully insert the needle into the target under teleoperated control.


\section{CONCLUSIONS}
We developed an 11-DOF redundant needle-biopsy robot that advances the previous 8-DOF system.
The new mechanical and electrical design improves workspace reachability in clinically relevant scenarios.
The proposed weighted Jacobian-based controller uses different weight policies to provide flexibility for scenarios requiring large-scale, fast motion or precise in-bore needle manipulation.
Additionally, null-space control enables subgoals—such as improving manipulability and maintaining the end-effector pose—enhancing adaptability in the confined bore while preserving dexterity for insertion and manipulation.
We validate our methods through extensive simulations and experiments and demonstrate the system in teleoperation.
\textcolor{black}{
Future work will integrate haptic feedback to guide the operator during insertion and compensate for respiratory motion and establish optimization framework to find the best weight policy for specific performance requirement.
}



\addtolength{\textheight}{-0cm}   






%

\bibliographystyle{IEEEtran}
\bibliography{citation}
\balance

\end{document}